\documentclass[lettersize,journal]{IEEEtran}
\usepackage{amsmath,amsfonts}
\usepackage{algorithmic}
\usepackage{algorithm}
\usepackage{indentfirst}
\usepackage{array}
\usepackage[caption=false,font=normalsize,labelfont=sf,textfont=sf]{subfig}
\usepackage{textcomp}
\usepackage{stfloats}
\usepackage{url}
\usepackage{verbatim}
\usepackage{graphicx}
\usepackage{cite}
\usepackage{stackengine}
\usepackage{textcomp}
\usepackage[utf8]{inputenc}
\begin{document}

\title{Accurate Tracking of Arabidopsis Root Cortex Cell Nuclei in 3D Time-Lapse Microscopy Images Based on Genetic Algorithm}

\author{Yu Song,  Tatsuaki Goh, Yinhao Li, Jiahua Dong, Shunsuke Miyashima, Yutaro Iwamoto, Yohei Kondo, Keiji Nakajima*, Yen-wei Chen*
\thanks{Yu Song and Tatsuaki Goh contributed equally to this work. Yen-wei Chen and Keiji Nakajima are the corresponding authors.}
\thanks{Y. Song, Y. Li and Y. Chen are with the College of Information Scinece and Engineering, Ritsumeikan University, 525-8577, Osaka, Japan (e-mail: yusong@fc.ritsumei.ac.jp, yin-li@fc.ritsumei.ac.jp, chen@is.ritsumei.ac.jp)}
\thanks{T. Goh and Keiji Nakajima are with the Graduate School of Science and Technology, Nara Institute of Science and Technology, 630-0192, Nara, Japan (e-mail:goh@bs.naist.jp, k-nakaji@bs.naist.jp)}
\thanks{S. Miyashima is with Bioresource Engineering Laboratory, Ishikawa Prefectural University, 921-8836, Ishiwaka, Japan (e-mail:s-miyash@bs.naist.jp)}
\thanks{Y. Iwamoto is with Department of Engineering Informatics, Osaka Electro-Communication University, 572-0833, Osaka, Japan (e-mail:yiwamoto@osakac.ac.jp)}
\thanks{Y. Kondo is with Exploratory Research Center on Life and Living Systems, National Institutes of Natural Sciences, 444-0867, Aichi, Japan (y-kondo@nibb.ac.jp)}
\thanks{J. Dong is with the College of Computer Science and Technology, Zhejiang University, 310027, Zhejiang, China (e-mail:djh20@zju.edu.cn)}}

\markboth{The paper is currently under review}%
{Shell \MakeLowercase{\textit{et al.}}: A Sample Article Using IEEEtran.cls for IEEE Journals}


\maketitle

\begin{abstract}
Arabidopsis is a widely used model plant to gain basic knowledge on plant physiology and development. Live imaging is an important technique to visualize and quantify elemental processes in plant development. To uncover novel theories underlying plant growth and cell division, accurate cell tracking on live imaging is of utmost importance. The commonly used cell tracking software, TrackMate, adopts tracking-by-detection fashion, which applies Laplacian of Gaussian (LoG) for blob detection, and Linear Assignment Problem (LAP) tracker for tracking. However, they do not perform sufficiently when cells are densely arranged. To alleviate the problems mentioned above, we propose an accurate tracking method based on Genetic algorithm (GA) using knowledge of Arabidopsis root cellular patterns and spatial relationship among volumes. Our method can be described as a coarse-to-fine method, in which we first conducted relatively easy line-level tracking of cell nuclei, then performed complicated nuclear tracking based on known linear arrangement of cell files and their spatial relationship between nuclei. Our method has been evaluated on a long-time live imaging dataset of Arabidopsis root tips, and with minor manual rectification, it accurately tracks nuclei. To the best of our knowledge, this research represents the first successful attempt to address a long-standing problem in the field of time-lapse microscopy in the root meristem by proposing an accurate tracking method for Arabidopsis root nuclei.
\end{abstract}


\section{Introduction}
\begin{figure}[h]%
	\centering
	\includegraphics[width=0.4\textwidth]{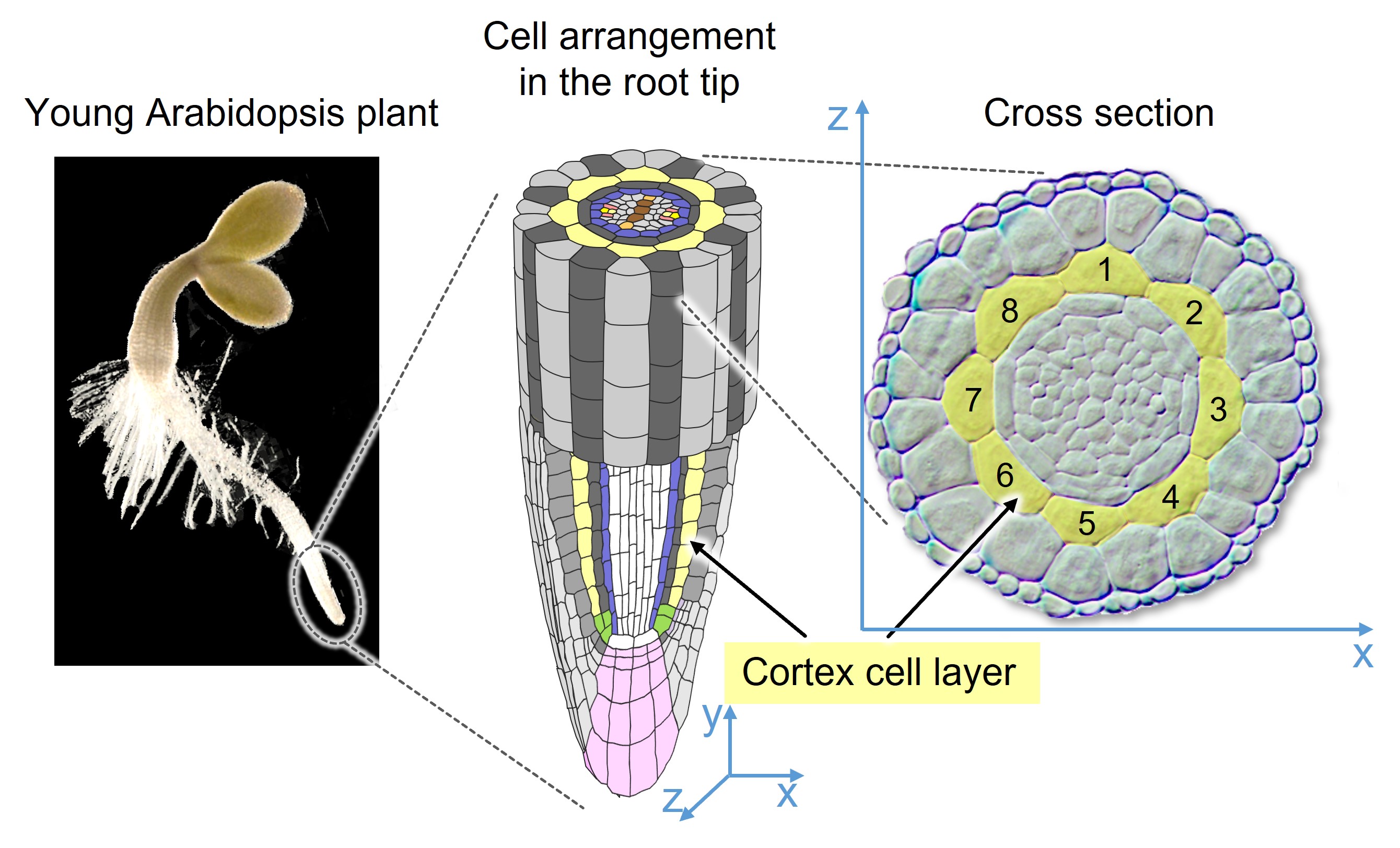}
	\caption{The Arabidopsis root image and the axial plane of Arabidopsis root, where the number of cortex nuclei is almost always eight.}\label{fig1}
\end{figure}

\IEEEPARstart{S}{tudies} using the roots of \textit{Arabidopsis thaliana} (Arabidopsis) plants have made a great contribution to establishing novel principles in plant development \cite{van1995cell, sabatini1999auxin, eldar2010functional, moreno2010oscillating}. As shown in Fig. \ref{fig1}, Arabidopsis roots have a simple tissue pattern and high optical transparency, and hence have been widely used as an ideal model system to study plant developmental processes. In order to follow their cellular dynamics, live imaging combined with mathematical modeling has become a key technology. Live cellular imaging technique has been used for the shoot apex for more than a decade \cite{heisler2005patterns, jonsson2006auxin}, but only recently has become available for the root apex by the introduction of automatic motion-tracking microscopes that can automatically follow a rapidly translocating root tip to keep it in the observation field \cite{Goh2022, rahni2019week, von2017live}. These motion-tracking microscopes with horizontal light axis are able to capture 3D time-lapse images (3D-space plus time-lapse) of Arabidopsis root tips. Among the radially arranged root tissue layers, the single cortex cell layer composed of eight cell lines (termed "cell files") is easily recognizable and thus used in this study to establish the algorithm to track the cell lineage by using fluorescently labelled images of cell nuclei as a proxy \cite{Goh2022}.\\

Cell-level progression of many plant developmental processes remains to be dissected. For example, it is still unknown how the division and elongation of thousands of root apical cells coordinately realize the dynamic root growth in response to environmental stimuli. To quantitatively dissect such complex processes, accurate tracking of cells and their lineages in the time axis is an essential technique. The root growth is driven by the sum of cell division and cell elongation. To understand the root growth dynamics, accurate tracking of cell division and cell elongation in relation to their lineages in time-lapse images is an essential task.\\

Several characteristics of Arabidopsis root cells make this tracking task challenging. Firstly, the Arabidopsis root cells are densely aligned along the lines, where spatial relationship (i.e., the distance between two consecutive  volumes of cells) cannot be used as a criterion to solve cell-cell associations. Secondly, long-term reliability   is required. During root growth, cell elongation and division occur frequently, and the live imaging is usually performed for a few days. These characteristics not only impose a requirement of accurate tracking between two consecutive frames, but also necessitates the tracking strategy to be stable enough for long-term observation. Thirdly, rotational movement of roots often makes it difficult to accurately track the cell files and the cell lineages. Last but not least, plant biologists need the quality of tracking results that is accurate enough to verify their hypothesis by mathematical simulation, where tracking accuracy should be as close to 100\% as possible.\\

To tackle the issues mentioned above, in this paper, we propose an accurate tracking method based on genetic algorithm (GA)\cite{mitchell1998introduction}. As shown in Fig. \ref{fig1}, based on the morphological regularity that Arabidopsis roots almost always have eight cortex cell files \cite{janes2018cellular}, the cortex cell files can be simplified into eight lines of their nuclei. Tracking each line of nuclei in the time space is considerably more practical than tracking respective nuclei. With the knowledge of line correspondence, nuclei from consecutive frames that belong to the same line can be easily tracked. To track each line of nuclei, we break the tracking problem into two parts: 1) separating eight lines of nuclei, and 2) link each line in the time space. Due to the rotational movement of roots, the shape of lines in different frames can curve considerably, making automatic separation of lines and their correlation across the time frames difficult, as is shown in Fig. \ref{fig4}. Our intuitive idea to solve this problem was to project nuclei onto the X-Z plane, as shown in Fig. \ref{fig1}, where the projected nuclei should be easily classified into eight lines. However, as we mentioned above, diversity of line shapes makes the projection fail in case lines of nuclei are not straight. Consequently, instead of projecting the nuclei on an X-Z plane, we needed a plane where projected nuclei can be explicitly separated into eight clusters.\\  

To solve the problem mentioned above, here we utilize GA to help find an optimal space for projection, which we will elaborate in the next section. We then use clustering method to automatically cluster the projected 2D nuclear locations into eight groups each of which represents a file of nuclei.  To track each line of nuclei in the time space, we utilize the spatial relationship between consecutive volumes. Specifically, for each line at a volume we extract the nuclei that are in the middle of the y axis (Definition of the coordinate is as shown in Fig. \ref{fig1}), so that we will have a total of eight nuclei in each frame arranged in an oval shape.  Then we utilize polar coordinate to track only one nucleus in the time space, after which remaining seven nuclei can be tracked easily since they are arranged in order by angle. \\

Our contribution is summarized as follows:\\

1. Unlike existing methods that rely on discrete optimization techniques such as Euclidean distance minimization—often failing in densely packed cell environments—we propose a two-step tracking framework. In Step 1, we leverage structural knowledge of Arabidopsis root cortex cells to perform line-level tracking, grouping cells into structured files. In Step 2, we apply nuclei-level tracking, integrating spatial relationships and cell type information for improved accuracy.\\

2. 	Our method explicitly optimizes the projection plane using a Genetic Algorithm (GA) to preserve the spatial structure of curved cell files before tracking. This biologically informed projection ensures alignment with the natural arrangement of Arabidopsis cortex cells, enabling automated clustering into cell groups prior to tracking. \\

3. Our method addresses the longstanding challenge of automatically tracking individual nuclei in time-lapse microscopy of root meristems, where nuclei are small and closely spaced. To the best of our knowledge, this is the first approach to achieve high-accuracy tracking of Arabidopsis root nuclei, which has already been adopted by biologists \cite{Goh2022}. Additionally, our method provides a reliable algorithm for generating ground-truth correspondences, supporting the development of advanced supervised AI models for nucleus tracking in plant biology research.

\section{Related Work}
\subsection{Cell tracking}
Cell tracking (or nuclear tracking in this study) can be viewed as a multi-object tracking problem, but with a minor difference in that cells usually exhibit similar appearance and divide during tracking. The conventional mainstream methods on muti-object tracking are tracking-by-detection methods. These methods can be summarized as a two-step fashion, detection and association. For detection, commonly used methods include conventional image processing methods like Laplacian of Gaussian (LoG) filter methods \cite{lindeberg1998feature}, anchor based deep learning-based methods such as Faster-RCNN \cite{ren2015faster}, Yolo methods \cite{redmon2018yolov3}, anchor-free methods such as Centernet \cite{zhou2019objects}, and Transformer-based methods such as DETR \cite{carion2020end}. For tracking strategies, most existing tracking methods often adopt discrete optimization methods, such as linear assignment problem (LAP) tracker \cite{jaqaman2008robust}, integer linear programming (ILP) \cite{bragantini2024large, jug2016moral}, IoU matching methods \cite{bewley2016simple}, Kalman Filter methods \cite{wang2019ranet, zhang2021fairmot}, LSTM methods \cite{milan2017online, kim2018multi} and Transformer-based methods \cite{meinhardt2022trackformer, sun2020transtrack}. Recently, some methods incorporated the association into detection model, in which detection and association are jointly performed at the same time, such as those published in \cite{bergmann2019tracking, chu2019famnet, feichtenhofer2017detect, hayashida2020mpm}. For biologists, a platform called Fiji/ImageJ and its plug-in program TrackMate are widely used to conduct cell tracking tasks \cite{schneider2012nih}.\\

Though the methods mentioned above give satisfactory results in some instances, it is insufficient in accurately tracking densely arranged Arabidopsis root tip cells. The LAP tracker algorithm used in the TrackMate works well in tracking particles undergoing Brownian motion, but fails when the cells are distributed densely and have blurry intercellular boundaries. For IoU mapping, since time-lapse images of Arabidopsis root tips are usually captured with a 30-min interval and include tropic movement of the roots and elongation of constituting cells, it may completely fail since temporal frames may have no intersection at all. The Kalman filter assumes that both the system and equations for the observation models are linear, but may fail in tracking Arabidopsis root cells, as cells divide and move with the tropic root growth. For optical-flow based method, the dense arrangement of root cells makes estimation of flow field extremely difficult. The LSTM and Transformer based methods often require a large amount of annotated data for training, which is a time-consuming and labor-intensive task or even not realistic in many cases due to the lack of sufficient annotated data for training. In this paper, we established a practical tracking method, which does not require annotated tracking trajectories, yet is able to achieve stable and accurate tracking of Arabidopsis root nuclei. 

\subsection{Genetic Algorithm}
Genetic algorithm (GA) mimics natural selection. It is commonly used to generate high-quality solutions to optimization and search problems \cite{mitchell1998introduction}. This method is used to solve many optimization problems in real cases \cite{chen2002estimating, zhu2022optimizing, capelli2017genetic, olmi2000genetic}. By generating offspring using biologically inspired operators like crossovers and mutations, optimal solutions for real world problems can be deduced if suitably encoded \cite{whitley1994genetic}. Two crucial components are required in a typical GA: 1) a genetic representation to encode the population, and 2) a fitness function to evaluate the solution domain. In GA, each individual in the population is represented as a binary string of 0s and 1s, or other forms of encodings \cite{whitley1994genetic}.  Usually, a population is randomly generated at the initial step. Since GA is an iterative process, each population within an iteration is also called a generation. Fitness functions evaluate the fitness of each individual in a population, where the individuals with better fitting are selected. Subsequently, the selected individual is modified through crossovers and mutations. The modified population serves as the starting population for the next iteration. 

\section{Proposed Method}
\subsection{Overview}
\begin{figure}[h]%
	\centering
	\captionsetup{justification=centering}
	\includegraphics[width=0.2\textwidth]{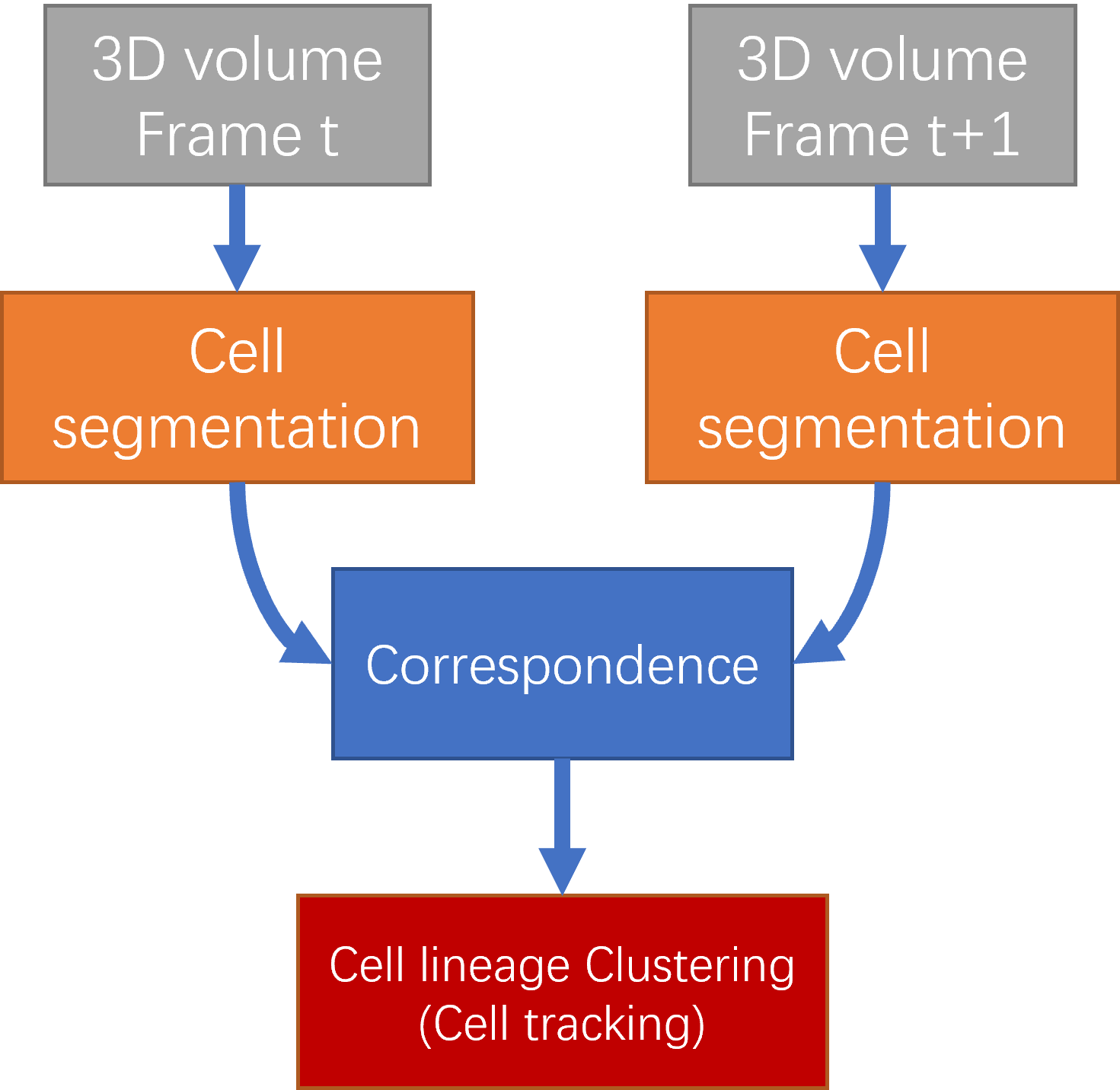}
	\caption{Our tracking-by-detection strategy. The cell segmentation method has been comprehensively detailed in a separate paper \cite{Goh2022}. In this paper, our primary focus is to provide an in-depth explanation of the tracking component.}\label{Overall}
\end{figure}

Our task is to perform nuclear tracking for Arabidopsis root cortex cells on 3D time-lapse mircoscopy images. We adopt the tracking-by-detection strategy as outlined in Fig. \ref{Overall}. For the detection step, we propose a deep learning-based segmentation method. While the detailed method is extensively explained in a separate paper \cite{Goh2022}, we provide a brief overview here.\\

We approach the detection problem as a segmentation task, where we perform three-class segmentation on 2D cell slices. We employ the UNet network \cite{RonnebergerFB15} for this purpose, taking 2D slices of volumes as inputs and generating three-class segmentation masks as outputs. These masks represent the background, non-dividing nuclei, and dividing nuclei. Our proposed method is supervised, as we have location coordinates and the classification of all nuclei (dividing or non-dividing) for every frame available through manual annotation by plant biologists. For the sake of simplicity in this paper, we directly utilize the manual annotations provided by the biologists.\\

\begin{figure}[!t]
	\centering
	\stackunder[5pt]{\includegraphics[width=0.17\textwidth]{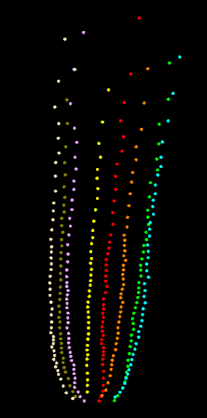}%
		\label{fig:a}}{frame=1, t=0}%
	\hfil
	\stackunder[5pt]{\includegraphics[width=0.17\textwidth]{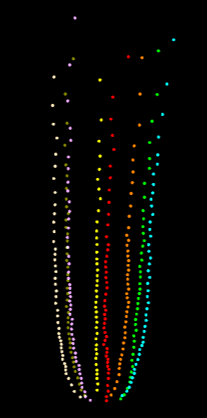}%
		\label{fig:b}}{frame=2, t=30min}%
	\hfil
	\stackunder[5pt]{\includegraphics[width=0.17\textwidth]{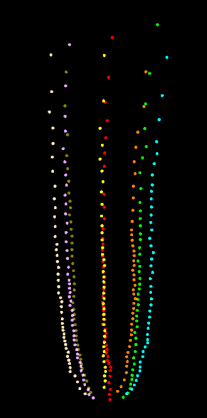}%
		\label{fig:c}}{frame=3, t=60min}%
	\hfil
	\stackunder[5pt]{\includegraphics[width=0.17\textwidth]{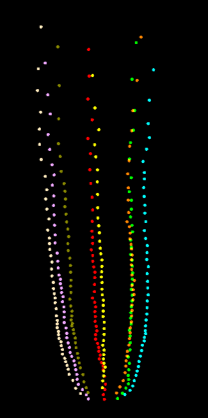}%
		\label{fig:d}}{frame=4, t=90min}%
	\caption{We show an example of consecutive volumes in the time space from frame 1 to frame 4 (representing volume at t=0,30min,60min,90min), where nuclei of each volume are represented by dots of eight colors and same color represent same lines of nuclei. Rotational movement of roots can be vividly observed in this figure.}
	\label{fig4}
\end{figure}

\begin{figure*}[!t]%
	\centering
	\captionsetup{justification=centering}
	\includegraphics[width=0.8\textwidth]{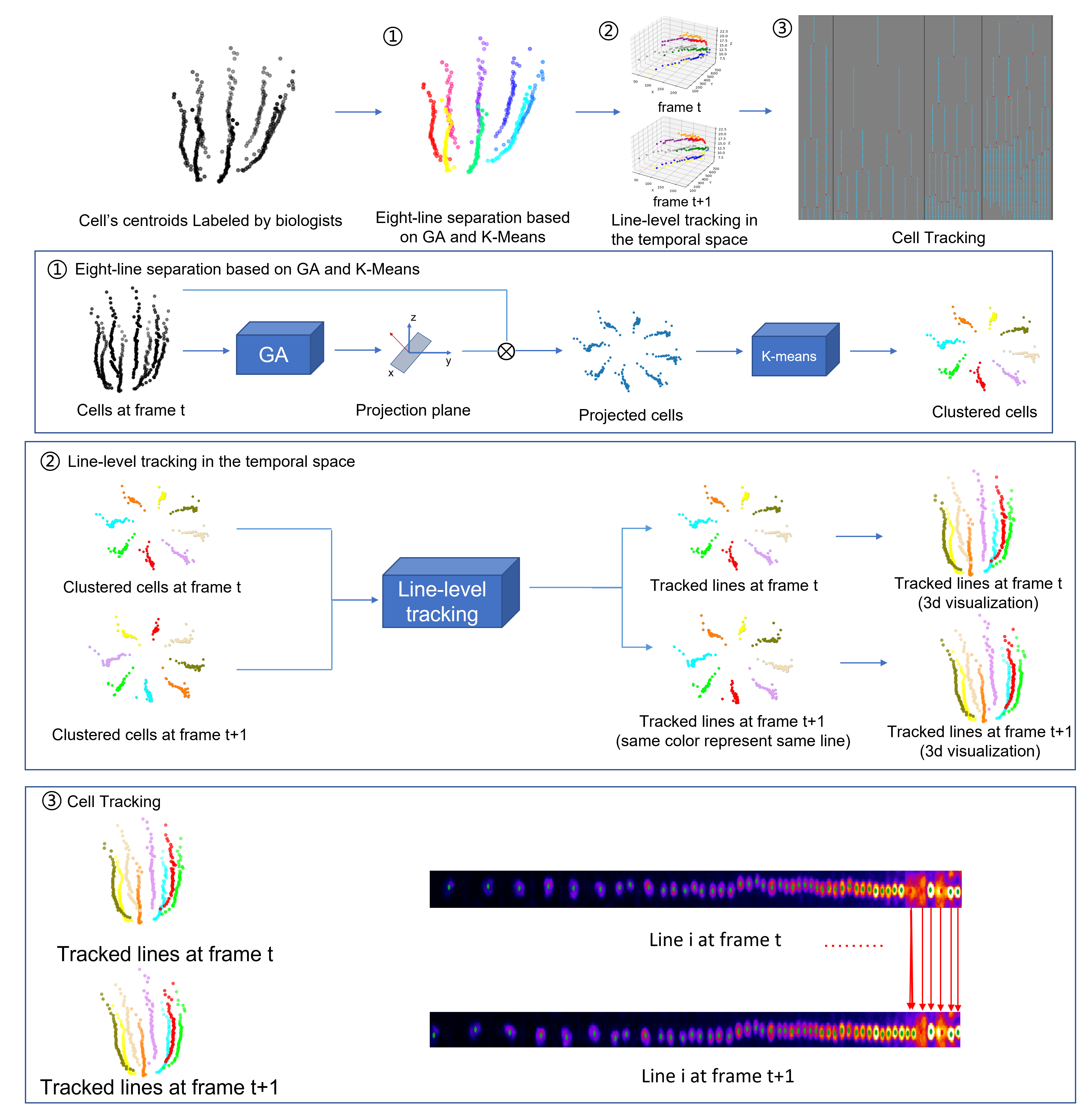}
	\caption{The overview of our proposed method.}\label{fig5}
\end{figure*}

In our tracking strategy, we utilize the tissue pattern information of the Arabidopsis root in which the cortical cells are organized into eight cell files and each cell contains a centrally localized nucleus. This allowed us to treat the 3D volume of cortical cell files at each time point as eight lines of nuclei. As to the method of automatically separating each volume to eight lines of nuclei, an intuitive idea would be to project all cells onto an X-Z plane, and then to apply a clustering method to cluster the projected nuclei into eight groups. However, as the root rotates and bends, cells are displaced, making the lines of nuclei curve in the temporal space instead of straight lines. As in the time-lapse images shown in Fig. \ref{fig4}, Arabidopsis roots can rotate in stochastic directions during the time-lapse imaging. In such cases, the projections of eight lines of nuclei on the X-Z plane will mix with each other, bringing difficulty in discriminating each line. In order to obtain easily separable projections, it is necessary to project the nuclei onto a space that facilities the separation of the eight groups. In order to solve this optimal problem, we here propose a Genetic Algorithm to find the suitable projection space. We will elaborate our GA in section \ref{secga}. After the projection, the projected nuclei are automatically clustered into eight groups each representing a line of nuclei. Up to this point, the eight lines of nuclei at each frame are already separated. Then, we correlate the lines between every two consecutive frames. We take one nucleus from the middle of each line for every frame. Based on the oval-shaped spatial relationship of selected eight nuclei, we utilize a polar coordinate to track one selected nucleus in the time space, as the remaining seven nuclei can be unambiguously tracked based on their regular arrangement by angle offset. After successfully tracking the lines in temporal space, we link the corresponding nuclei of two consecutive volumes based on the tracked lines and the nucleus type (i.e. dividing or non-dividing). The overall architecture of our proposed method is summarized in Fig. \ref{fig5}.

\subsection{Genetic Algorithm}\label{secga}
\begin{figure}[h]%
	\centering
	\captionsetup{justification=centering}
	\includegraphics[width=0.4\textwidth]{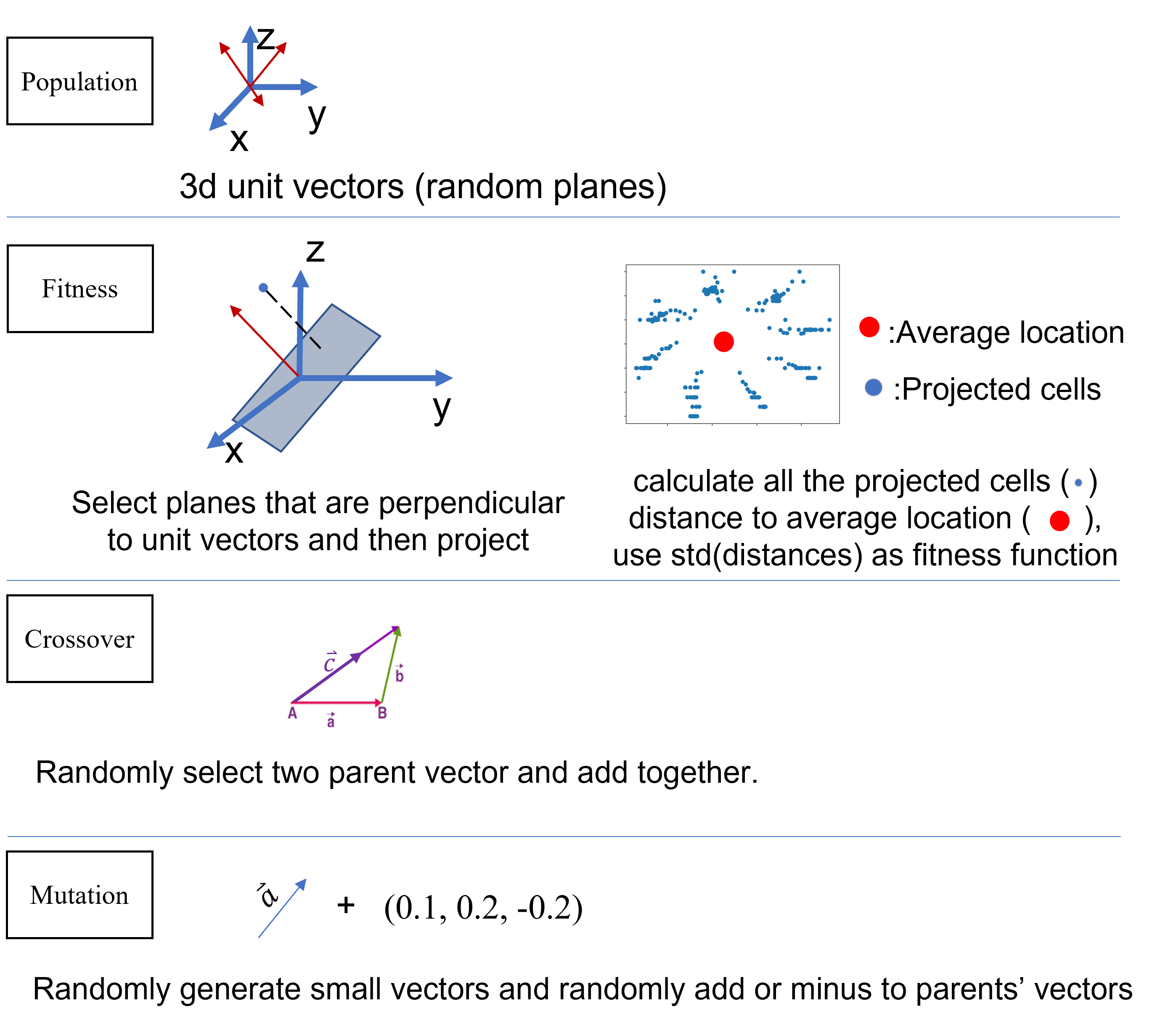}
	\caption{The main components of the proposed GA.}\label{fig6}
\end{figure}

Our proposed GA aims to find an optimal space where different lines of nuclei can be easily separated. The input of the GA are 3D coordinates of nuclei at time $t$: $X_I^t \in R^{I * 3}$, where $I$ represents the number of nuclei cells at time $t$. The 'chromosome' in our GA is randomly generated projection spaces. Through the selection using a defined 'fitness' functions followed by a 'mating' behavior, the next generation is generated. After many iterations, our proposed GA generates the final group. We select the 'chromosome' with best fitness in the final group as the optimized projection space. The main components are shown in Fig. \ref{fig6}.  \\

\subsubsection{Chromosome}
In order to use GA to solve an optimization problem, encoding of individual is of vital importance. Usually, individuals are encoded into strings of 0 and 1. Such individuals can also be named as chromosomes. The chromosome in our task is a plane for projection. To specify a plane, we can use a normal vector and a point location on the plane. With the chosen point on the plane and the normal vector, the plane can be represented using a vector equation. The vector equation for a plane is often written as: $P = P_0 + t * N$, where $P$ represents a generic point on the plane, $P_0$ represents the selected point on the plane, $N$ represents the normal vector to the plane and $t$ represents a parameter that can vary over real numbers. In our task, we use a unit vector $\vec{a} = (x, y, z)$ as $N$, coordinate origin $(0, 0, 0)$ as $P_0$ to represent a plane, a setting slightly different from those used in conventional GA encodings. Thus, the generic representation of our chromosome can be represented as $\vec{a} = (x, y, z)$. \\

\subsubsection{Initialization}
The size of the initial population depends usually on the nature of the problem. It usually ranges from hundreds to thousands of individuals. In our task, we randomly generated a thousand unit vectors as the initialization, where each unit vector represents a projection plane. \\

\subsubsection{Selection}
For population in each iteration, a small portion will be selected based on the defined fitness function. The more fitted individuals will have the larger chance to be selected. Fitness function aims to evaluate each individual and measures the quality of the represented groups. Usually, fitness function is defined according to the specific problem of interest. The appropriate fitness function will play a key role in the performance of GA. \\

In our task, we aim to find a plane for projection, where the projected nuclei will be easily separated into eight groups. In other words, the projected nuclei for each line should be close enough with each other to be clustered together. Therefore, for the projected nuclei, we calculate the distance of each nucleus to the centroid of the projected nuclei. If the eight clusters are easily recognized, the distances from the center location to each nucleus is similar. That is to say, the standard deviation of the distances will be substantially small. Based on such analysis, our fitness function is defined as the standard deviation of the distances from projected nuclei to the center of projected nuclei, which can be described as Eq. (\ref{std}), where $N$ represents the number of nuclei, $x_i$ represents the distance between nucleus $i$ and the center of the projected nuclei, and $\overline{x}$ represents the averaged distances $x_i, i=1, 2, ..., N$.\\

\begin{equation}
	fitness = \sqrt{\frac{1}{N-1} \sum_{i=1}^N (x_i - \overline{x})^2}
	\label{std}
\end{equation}

 We present an example of fitness calculation in Figure \ref{figFit}. In Figure \ref{figFit}, (a) illustrates nuclei projected onto the x-z plane, while (b) showcases nuclei projected onto the plane determined by the Genetic Algorithm (GA). In (c) and (d), we can observe the distances of each projected nucleus from the average position of the projected nuclei. Even through visual inspection, it is evident that the standard deviation of the lines in (c) is significantly greater than that in (d), not to mention the quantitative results obtained from the fitness calculation in both scenarios. Consequently, the fitness function continues to select the appropriate projection plane until the termination condition is satisfied.\\

\subsubsection{Crossover and Mutation}
The next step is to generate the next generation based on the selection by the fitness function. This step includes two operators: crossover and mutation. Usually, a pair of 'parent' solutions will be selected and used to produce a 'child’ solution after the application of either crossover or mutation. The created 'child' solution will typically share some features with those of the parents. This procedure will be repeated  until the number of generations meets the need for the next iteration. In our cases, the chromosome is represented as a unit vector at the coordinate origin, which is not encoded as 0s and 1s, therefore, the implementation of crossover and mutation also are different from those used for chromosomes encoded as 0s and 1s.\\ 

For crossover, we select two 'parent' chromosomes, and add them together, then we take the normal vector of the added vector as a 'child' vector. For mutation, we randomly generate small vectors, then randomly add or substract them to selected vectors. We repeat this procedure until the number of children meets the number of sets for the next generation.\\

\subsubsection{Termination}
There are many ways to terminate the GA. For example, we can set a minimum criterion until a solution meets. In our task, if the solution reaches a plateau and no longer creates more fitted solutions, we will terminate the GA. Specifically, We set the max iteration number as 300 and the tolerance as $1e^{-4}$. \\

\subsubsection{Iteration}
For each generation, fitness function is applied to individuals. Among them, more fitted individuals will be selected. After crossover and mutation, they will form the next generation. This process will be repeated until the termination condition is met.

\subsection{K-means Clustering}
Using GA, we already project nuclei in a given 3D volume to an optimized plane for separation. Next task is to automatically separate the projected nuclei into eight groups, with each group represented as a line of nuclei. Here, we adopt the K-means clustering method. K-means clustering aims to partition n observations into k clusters in which each observation is clustered with the nearest mean (a cluster center or a cluster centroid), giving rise to a prototype of the cluster. The algorithm as described by \cite{macqueen1967some} starts with a random set of $k$ center-points ($\mu$). During each update step, all observations x are assigned to their nearest center-point (see equation \ref{eqn:kmeans_assign_step}). In the standard algorithm, each observation can be assignment to a sole center. If multiple centers have the same distance to the observation, a randomly chosen center is assigned, \\

\begin{equation}
	S_i^{(t)} = \big \{ x_p : \big \| x_p - \mu^{(t)}_i \big \|^2 \le \big \| x_p - \mu^{(t)}_j \big \|^2 \ \forall j, 1 \le j \le k \big\}
	\label{eqn:kmeans_assign_step}
\end{equation}

where $S_i$ is the $i_{th}$ set, $\mu_i$ is the mean of points of $S_i$, each $x_{p}$ is assigned to one $S^{(t)}$. Subsequently, the center-points are repositioned by calculating the mean of the assigned observations to the respective center-points, as shown in Eq. \ref{eqn:kmeans_update_step}.

\begin{equation}
	\mu^{(t+1)}_i = \frac{1}{|S^{(t)}_i|} \sum_{x_j \in S^{(t)}_i} x_j
	\label{eqn:kmeans_update_step}
\end{equation}

The update process is repeated until all observations remain assigned to the previously assigned center-points and therefore the center-points would not be updated anymore.\\

In this step, we use K-means to separate our projected nuclei into eight clusters. Thus, each cluster is a line of nuclei. We apply this to every time frame of our data, so that each volume can now be viewed as eight lines of nuclei. \\

\subsection{Manual Refinement}
\begin{figure}[h]%
	\centering
	\captionsetup{justification=centering}
	\includegraphics[width=0.3\textwidth]{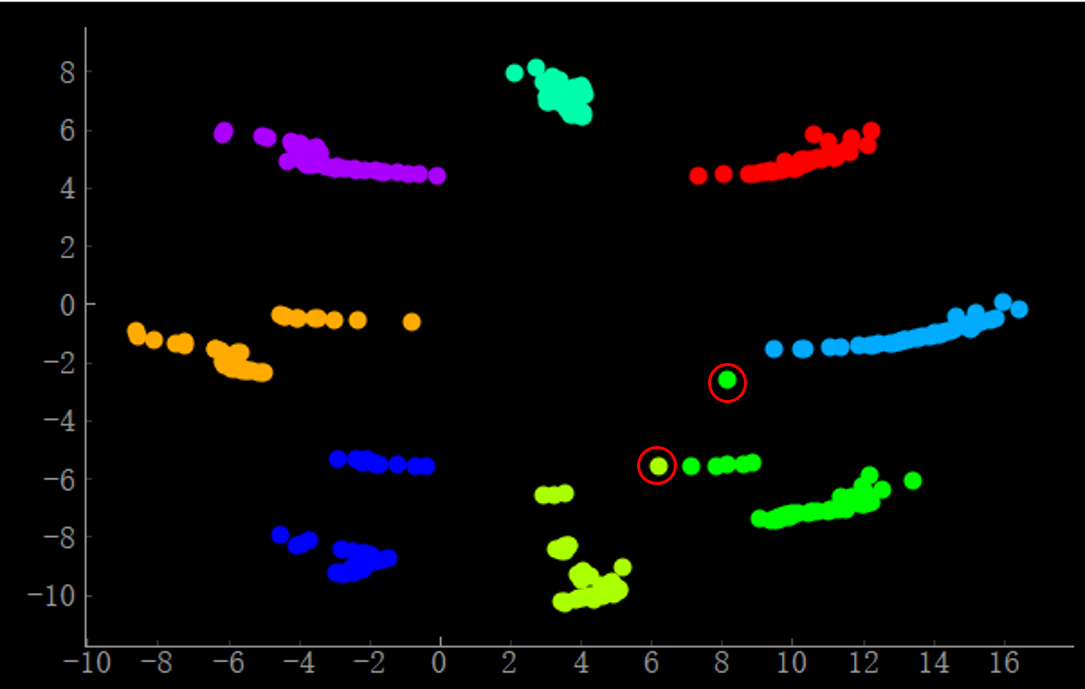}
	\caption{An example of K-means clustering result, where each color represents a cluster. Red circle represents k-means clustering errors.}
	\label{k-means}
\end{figure}

Even with the use of the projection space selected by our proposed GA, K-means clustering may still generate errors. An example of k-means results is shown in Fig. \ref{k-means}, where the same color represents the same cluster. The nuclei circled in red are likely to be errors that need to be refined. Here, we manually rectify these errors. 

\subsection{Line Correspondence}
\begin{figure}[h]%
	\centering
	\captionsetup{justification=centering}
	\includegraphics[width=0.35\textwidth]{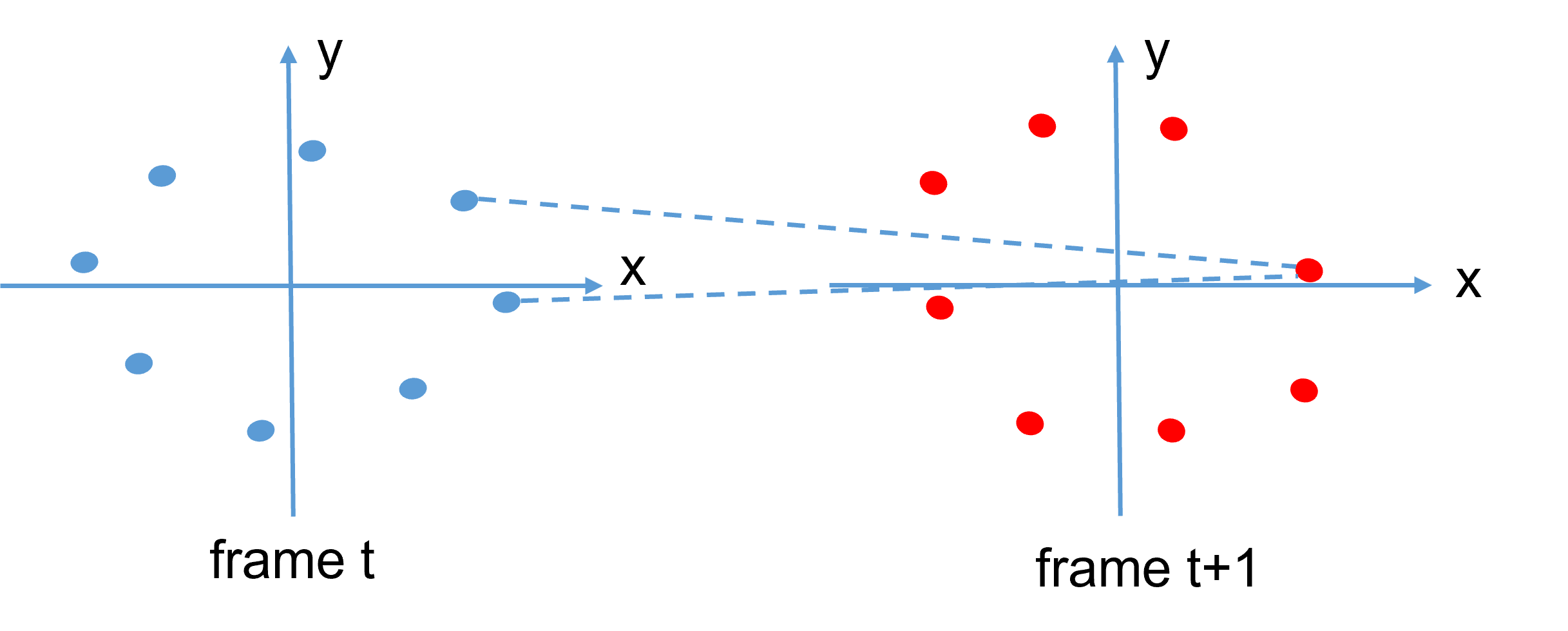}
	\caption{An example of corresponding lines between frame $t$ and frame $t+1$ in Cartesian coordinate through smallest Euclidean distance, where cases that two nuclei in frame $t$ having a same correspondence in frame $t+1$ occur (blue dash lines).}\label{cartesian_eg}
\end{figure}

\begin{figure}[!t]
	\centering
	\subfloat[]{\includegraphics[width=0.2\textwidth]{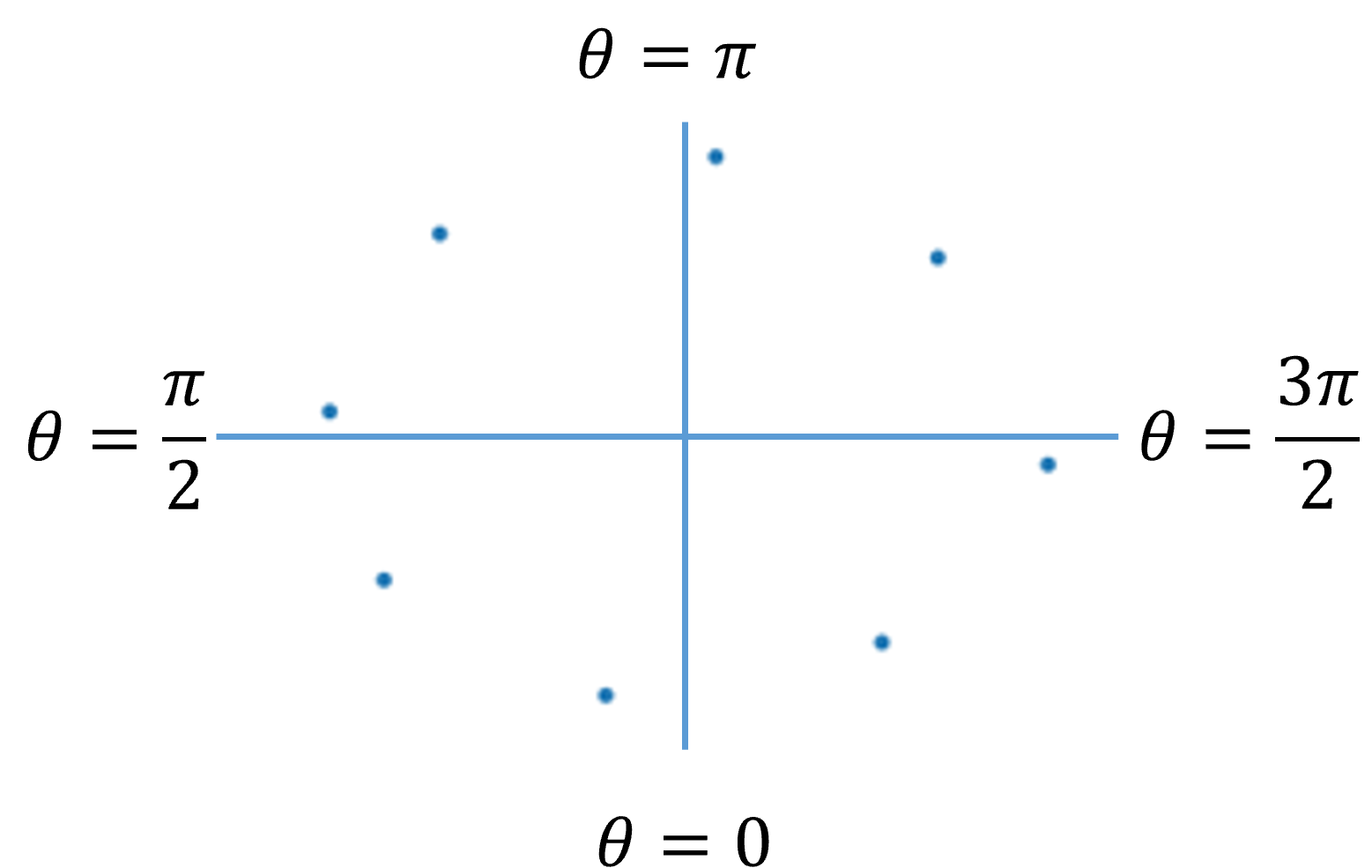}%
		\label{fig:line_a}}
	\hfil
	\subfloat[]{\includegraphics[width=0.25\textwidth]{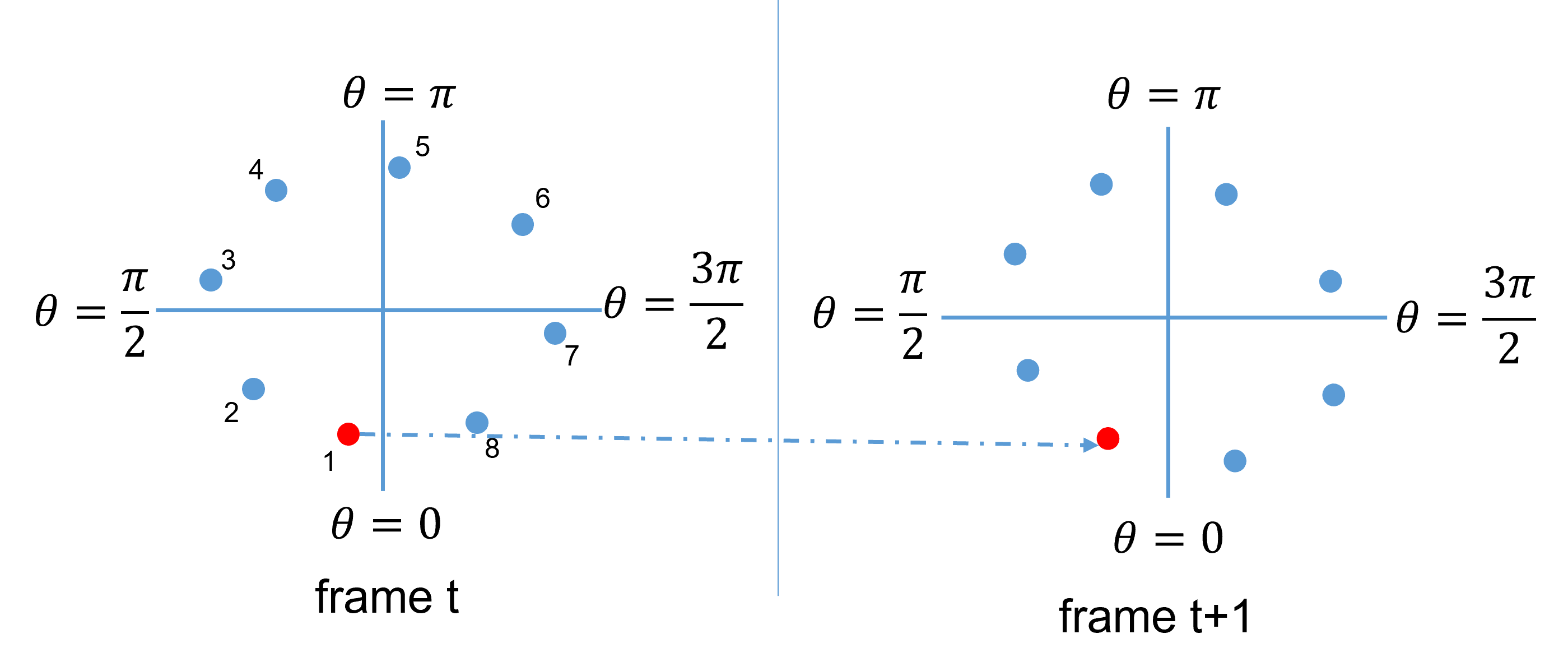}%
		\label{fig:line_b}}
	\caption{We use location information and use polar coordinates to track each of 8 lines in the time axis. (a) The axial plane of 8 nuclei in the middle of y axis, where each cell belongs to a different line. (b)Using polar coordinates for tracking each line.}
	\label{fig7}
\end{figure}

After partitioning the nuclei in each volume into eight lines, we need to link each line between two consecutive volumes, namely, to track each line along the time axis. The inputs in this step are the clustered eight lines of nuclei from two consecutive frames, which can be represented as ($m_i^t \in R^{8 * 3}, m_i^{t+1} \in R^{8 * 3}$),  where $i$ represents the line index ranging from 1 to 8, $t$ represents the time frame, and each line $m$ is denoted by a single nucleus located in the middle of that line. For the image of every volume, we take one nucleus from each line in the middle of the y axis in the original image space, as shown in Fig. \ref{fig:line_a}. Then, we utilize the coordinate information of the eight lines to link two consecutive volumes. Our initial aim of correlating corresponding lines by Euclidean distance resulted in the disruption of line order. As shown in Fig. \ref{cartesian_eg}, using cartesian coordinate and link nuclei through smallest Euclidean distance, there are cases where two nuclei in frame $t$ may have a same correspondence in frame $t+1$. Additionally, for substantial displacements, the Euclidean distance in Cartesian coordinates often proves inadequate. Thus, we do not adopt the shortest Euclidean distance, but instead, track only one of the eight nuclei, since the eight nuclei are arranged in an oval shape and thereby allowing the remaining seven nuclei to be correlated according to the angle information as shown in Fig. \ref{fig:line_b}. To do this, we transfer the cartesian coordinate of the eight nuclei to polar coordinates, and link the nuclei with the smallest positive angle between frame $t+ 1$ and frame $t$. Once the nucleus we selected is tracked between two consecutive frames, the other nuclei are therefore able to be unambiguously correlated using the angle information. We repeat this process from the second volume to the last volume. Now, all eight lines are tracked along the time axis. \\

It is worth noting that our dataset has a time interval of 30 minutes between two consecutive frames, a parameter determined by biologists. Illustrations of consecutive frames from our dataset are presented in Fig .\ref{fig4}. These visual representations in Fig .\ref{fig4} highlight that the angular displacement of each nucleus line is relatively small in comparison to the Euclidean distances between individual nuclei. If the time interval were significantly larger, the angular displacement between two consecutive frames could become too substantial, rendering the tracking of each line unfeasible.\\

In contrast to software such as Trackmate in ImageJ, our proposed method tracks each line of nuclei rather than each nucleus, based on their rotation degree in a polar coordinate system. This decision is based on our observation of minimal angular deviations between consecutive frames. We believe this choice is one of the key factors contributing to the success of our proposed tracking system. \\

\subsection{Nucleus Tracking (Nuclear lineage Clustering)}
Once the lines of nuclei are precisely tracked, tracking of each nucleus can be easily implemented by incorporating the information of the nucleus type (non-mitotic or mitotic). The inputs are the nuclei in consecutive frame $t$ and $t+1$: $X_i^t \in R^{I * 3}, X_j^{t+1} \in R^{J * 3}$, where $I$ represents number of nuclei at time $t$, J represents the number of nuclei at time $t+1$. For $l_{th}$ line in the two consecutive volumes $t$ and $t+1$, we link every corresponding nuclei from the bottom (root tip) to the top (root base) along the y axis, where y axis is shown as Fig. \ref{fig1}. For a nucleus of $l_{th}$ line in frame $t$, if the type of the nuclei is non-mitotic, then it is linked to one nucleus of $l_{th}$ line in volume $t+1$. If the type is mitotic, namely the nucleus is dividing into two, the corresponding two nuclei of line $l$ in volume $t+1$  were regarded as the daughter nuclei of the nucleus in volume $t$, if the two nuclei are non-mitotic (i.e. completing the division). We applied this correlation between every pair of two consecutive volumes. The nucleus tracking procedure is described as alg. \ref{alg1}. In the algorithm, $i \leq 8$ means that there are eight lines in every volume, and $argsort(l^t [:, :, 2])$ aims to rank nuclei by y coordinate in a descending order.

\section{Experiments}
\subsection{Dataset}
\begin{figure}[!t]%
	\centering
	\captionsetup{justification=centering}
	\includegraphics[width=0.3\textwidth]{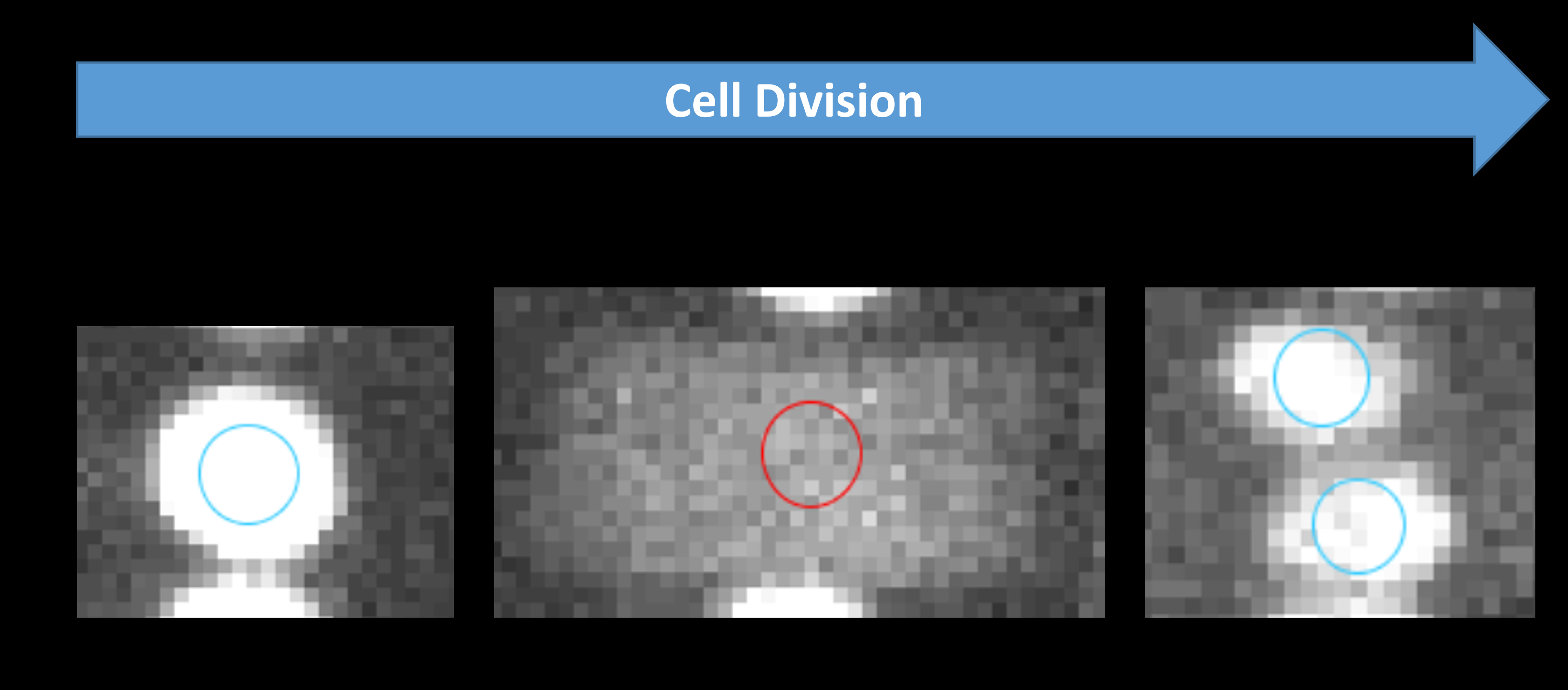}
	\caption{Cell division process: a non-dividing nucleus (left labeled by blue circle) becomes a dividing nucleus (middle labeled by red circle), finally divides to two daughter nuclei (right labeled by blue circle).}\label{fig8}
\end{figure}

The datasets used in this experiment were 3D time-lapse images of Arabidopsis roots whose cortex cell nuclei are fluorescently labelled by Red Fluorescent Protein (RFP), taken by a custom-made motion-tracking microscope \cite{Goh2022}. We used three data sets with the size of $101(t) * 27(slice) * 1000(H) * 454 (W)$, $144(t) * 29(slice) * 1267(H) * 749 (W)$, and $50(t) * 29(slice) * 1054(H) * 521 (W)$. Their intervals between two consecutive frames were 30 minutes, 30 minutes, and 15 minutes, respectively. The size of a Voxel was $2.5 * 0.61 * 0.61 \mu m^3$. Location coordinates and types of nuclei were manually annotated by an author expertizing plant developmental biology using the ImageJ software. The type of nuclei were mitotic and non-mitotic, which are those dividing to two daughter nuclei and those not dividing, respectively. Representative nuclear division images are shown in Fig. \ref{fig8}, where a non-mitotic nucleus transitions to a mitotic nucleus and then becomes two nuclei in the temporal space. We labelled nuclei in all 144 time-frame datasets and use them to evaluate the feasibility of the proposed method.

\begin{algorithm}
	\caption{Cell Tracking Algorithm}
	\begin{algorithmic}[1]
		\REQUIRE Tracked lines, Cell location and type \\
		\hspace*{\algorithmicindent} \textbf{Input:} {$i$-th line $l_i$ of nuclei at frame $t$: $l_i^t \in \mathcal{R}^{m}$, $i$-th line $l_i$ of nuclei at frame $t+1$: $l_i^{t+1} \in \mathcal{R}^{n}$} \\
		\hspace*{\algorithmicindent} \textbf{Output:} {Cell’s correspondence between $l_i^t$ and $l_i^{t+1}$} 
		\WHILE{$i \leq 8$}
		\STATE  $l_i^t = argsort(l_i^t [:, :, 2])$,  $ l_i^{t+1} = argsort(l_i^{t+1} [:, :, 2])$, $j=1$, $k=1$
		\COMMENT{sort nuclei cells by y coordinates in a descending manner}
		\WHILE{$j \leq m$, $k \leq n$}
		\IF{$l_i^t[j]$ is not mitotic}
		\STATE $ l_i^t[j].correspondence = l_i^{t+1}[k]$
		\STATE $j += 1, k += 1$
		\COMMENT{For non-dividing nuclei cells, we correspond one cell in $l_i^t$ to one cell in $l_i^{t+1}$.}
		\ELSE
		\STATE $ l_i^t[j].correspondence =  [l_i^{t+1}[k], l_i^{t+1}[k+1]]$
		\STATE $j += 1, k += 2$
		\COMMENT{For dividing nuclei cells,  we correspond one cell in $l_i^t$ to two daughter cells in $l_i^{t+1}$.}
		\ENDIF
		\ENDWHILE
		\STATE $i += 1$
		\ENDWHILE
	\end{algorithmic}
	\label{alg1}
\end{algorithm}

\subsection{Parameter Settings}

We set the following parameters in our GA: population: 1000, maximum iteration: 200, Crossover probability: 0.8, mutation probability: 0.1. In terms of the K-means clustering, the parameters are set as follows: number of cluster: 8, number of time K-means will be run with different centroids: 10, max number of iterations: 300, tolerance: $1e^{-4}$.

\subsection{Results}

\begin{figure}[!t]
	\centering
	\subfloat[]{\includegraphics[width=0.16\textwidth]{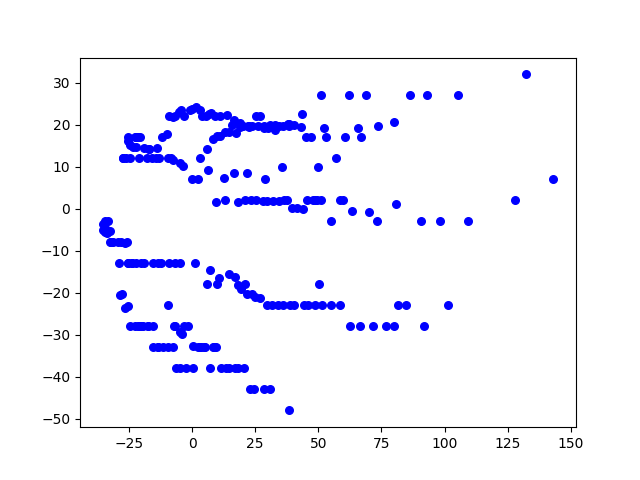}%
		\label{fig:x_z}}
	\hfil
		\subfloat[]{\includegraphics[width=0.16\textwidth]{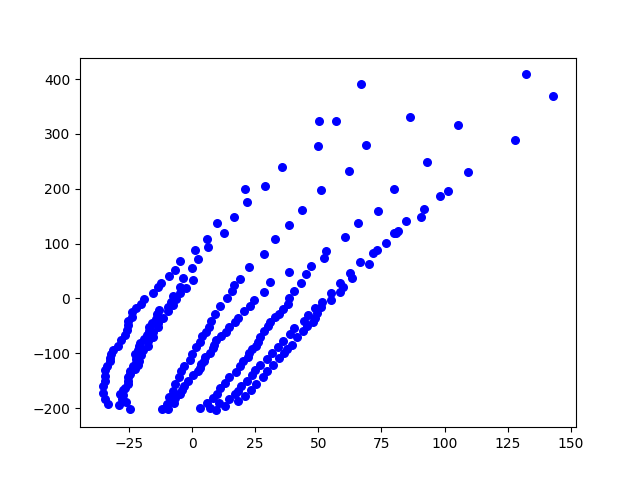}%
		\label{fig:x_y}}
	\hfil
		\subfloat[]{\includegraphics[width=0.16\textwidth]{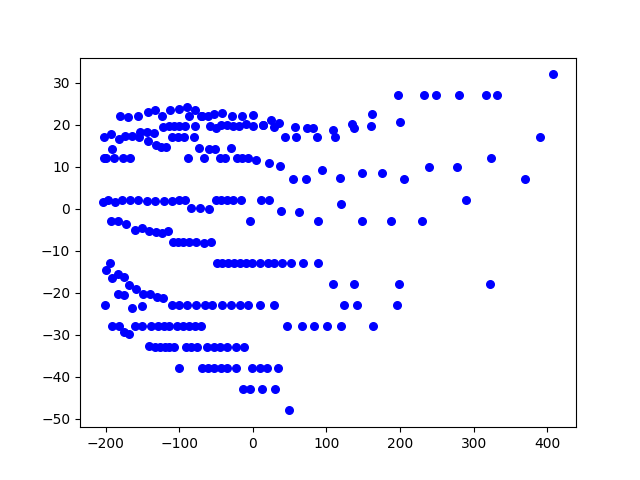}%
		\label{fig:y_z}}
	\hfil
	\subfloat[]{\includegraphics[width=0.16\textwidth]{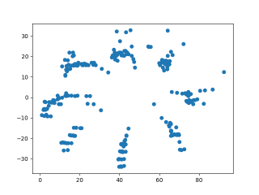}%
		\label{fig:ga_result}}
	\hfil
	\subfloat[]{\includegraphics[width=0.16\textwidth]{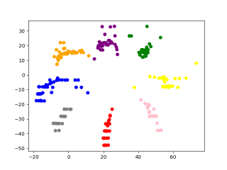}%
		\label{fig:cluster}}
	\hfil
	\subfloat[]{\includegraphics[width=0.16\textwidth]{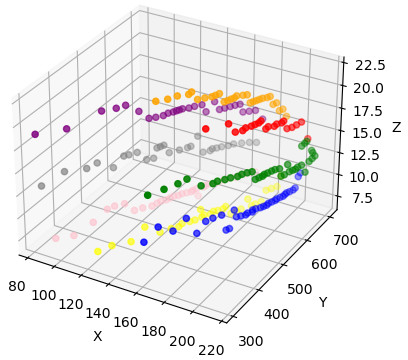}%
		\label{fig:3d}}
	\caption{Result examples. (a)The projection to x-z plane. (b)The projection to x-y plane. (c)The projection to y-z plane. (d)The projection using plane selected by GA. (e)K-means clustering result on (d). (f)3D plot of 8 lines of nuclei.}
	\label{fig9}
\end{figure}

\begin{figure}[!t]
	\centering
	
	\stackunder[5pt]{\includegraphics[width=0.2\textwidth]{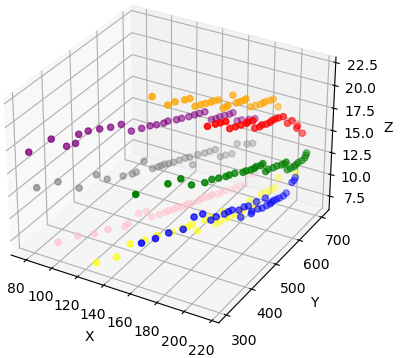}%
		\label{fig:0}}{frame=1, t=0}%
	\hfil
	\stackunder[5pt]{\includegraphics[width=0.2\textwidth]{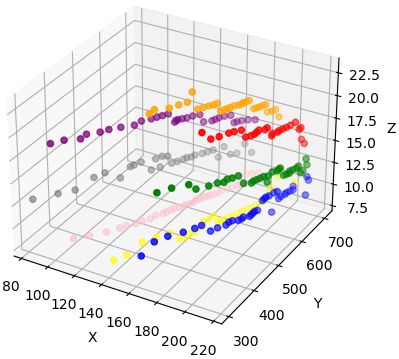}%
		\label{fig:1}}{frame=2, t=30min}%
	\hfil
	\stackunder[5pt]{\includegraphics[width=0.2\textwidth]{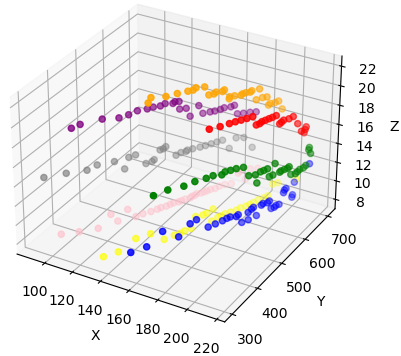}%
		\label{fig:2}}{frame=3, t=60min}%
	\hfil
	\stackunder[5pt]{\includegraphics[width=0.2\textwidth]{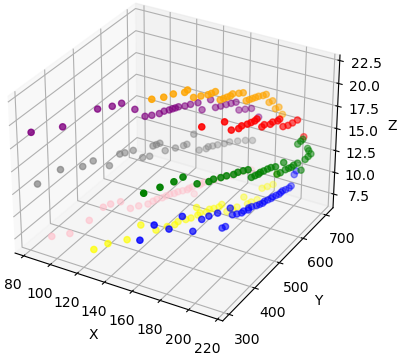}%
		\label{fig:3}}{frame=4, t=90min}%
	\caption{Line correspondence result from frame=1 to frame=4, where 8 line is represented in unique color and same color represents same line.}
	\label{fig10}
\end{figure}

\begin{figure}[h]
	\centering
	\subfloat{\includegraphics[width=0.5\textwidth]{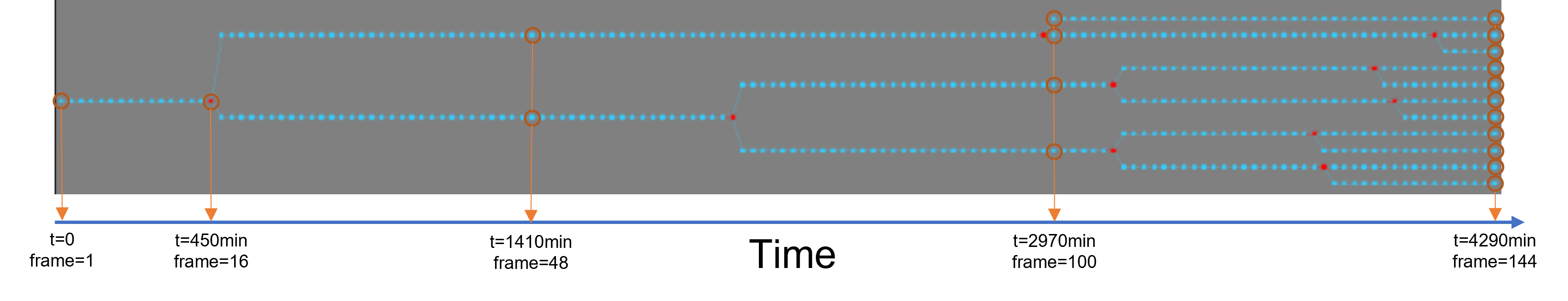}%
		\label{fig:spt1}}
	\hfil
	\subfloat{\includegraphics[width=0.5\textwidth]{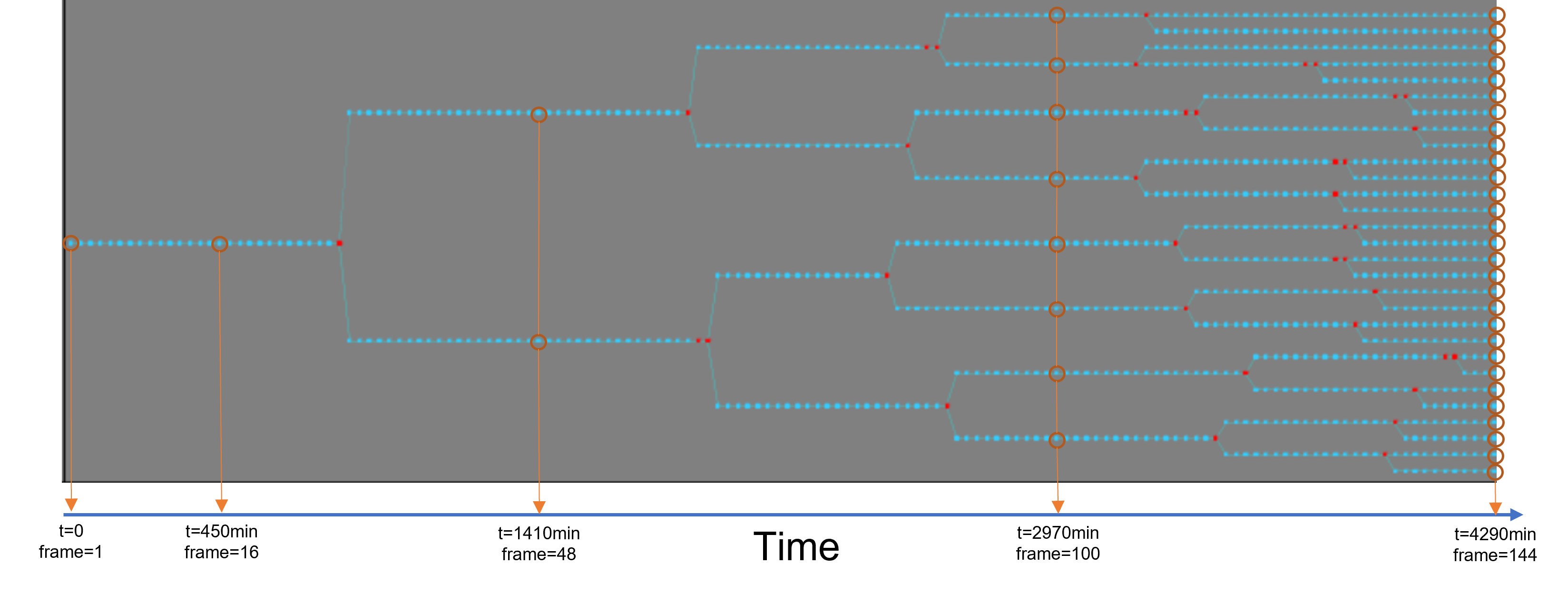}%
		\label{fig:spt2}}
	\caption{Example of two cells tracked in 144 time-frame. In this figure, x axis means time space, red dots represent mitotic nuclei and blue dots represent non-mitotic nuclei. }
	\label{figSPT}
\end{figure}

\begin{figure}[h]%
	\centering
	\captionsetup{justification=centering}
	\includegraphics[width=0.48\textwidth]{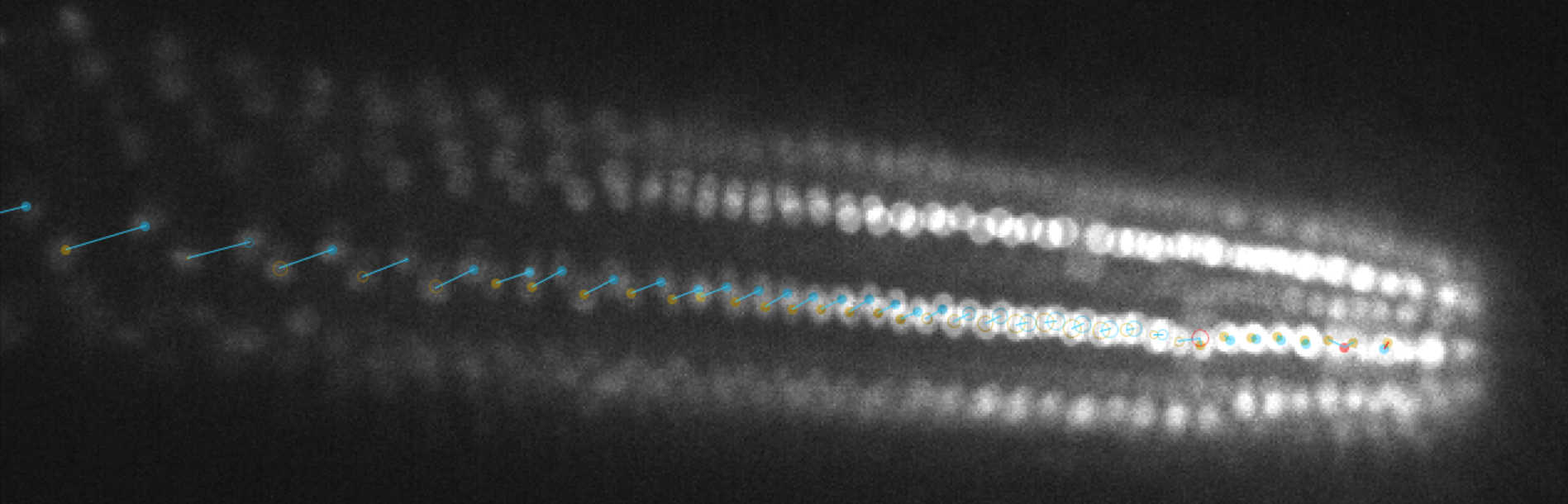}
	\caption{An example of tracked nuclei, we blend two consecutive frames together, where circle means nuclei and line means nuclei' correspondence. Blue circle presents nuclei at frame $t-1$, yellow circle represents nuclei as frame $t$, red circle represents mitotic nuclei. }\label{figtracked}
\end{figure}

An example of our result is shown in Fig. \ref{fig:ga_result}. Compared with a direct projection to an X-Z plane, an X-Y plane and a Y-Z plane shown in Fig. \ref{fig:x_z}, Fig. \ref{fig:x_y} and Fig. \ref{fig:y_z} respectively, the plane selected by our proposed GA allowed better partition of projected nuclei into eight groups. After the K-means clustering, the eight groups representing the eight lines of nuclei are preliminary clustered as shown in Fig.  \ref{fig:cluster}. As shown in Fig. \ref{fig:3d}, presentation of the clustering result back in the original 3D space clearly visualized correct separation of the eight lines of nuclei. After successfully separating each volume into eight lines of nuclei, we conducted line-based tracking between successive time frames. As shown in Fig. \ref{fig10} where the dots in the same color represent corresponding lines across the time frames, eight lines of nuclei were successfully tracked in the time axis. Finally, based on the line-correspondence result and nuclear type information, the tracking task was successfully completed as shown in the examples of two cells tracked in 144 time-frame in Fig. \ref{figSPT}. With the data representation shown in Fig. \ref{figSPT}, lineage information of the cells at the onset of time-lapse imaging can be easily recognized. An example of tracked nuclei overlaid on two consecutive slice images is shown in Fig. \ref{figtracked}, where all nuclei were tracked correctly, including those divided to give two daughter nuclei.

\subsection{Comparison with existing methods}\label{subsec10}
TrackMate is an open-source software that has been widely used by biologists. It provides tracking-by-detection tools for cell tracking tasks. For detection, it uses a LoG (Laplacian of Gaussian) or a DoG (Difference of Gaussians) filter to detect cell blobs. For tracking, it uses the LAP tracker (Linear Assignment Problem) method. The provided tracking-by-detection method works effectively in cases where particles undergo Brownian motion, which makes TrackMate a preferred software for biological applications. Therefore, we compared our method with the LAP tracker used in TrackMate. For cell detection, we used annotations made by biologists to ensure the same prerequisite with our experiments. We use True Positive (TP), False Positive (FP), False Negative (FN), and Precision and Recall for evaluation. If a predicted link between two nuclei exists in the ground-truth, we consider it as a TP. If a predicted link between two nuclei does not exist in the ground-truth, we consider it as a FP. If a link in the ground-truth does not exist in the tracking results, we consider it as a FN.

\begin{table*}[!t]
	\caption{Performance comparison between ImageJ and proposed methods on 144 frames\label{tab:tab2}}
	\centering
	\begin{tabular}{@{}llllll@{}}
		\hline
		& True Positive  & False Negative & False Positive & Precision & Recall \\
		\hline
		LAP tracker (TrackMate) & 26223   & 24752 & 21632 &54.8\% &51.4\% \\
		Ours (w/o manual refinement on clustering result) & 36324   & 14651 &6618 &84.6\%  &71.3\%   \\
		Ours &50975 &0 &0 &100\%  &100\% \\
		\hline
	\end{tabular}
	\footnotetext{The comparison of clustering accuracy between GA and PCA}
\end{table*}

 \hfill \break
 As shown in Table \ref{tab:tab2}, even without manual refinement of K-means clustering, where some nuclei remained clustered in wrong lines, our result achieved 84.6\% precision and 71.3\% recall,  as compared with 54.8\% and 51.4\%, respectively, achieved by TrackMate. Thus, even using uncorrected line correspondence, our tracking already outperformed that by LAP tracker. With manual refinement of minor clustering errors, our method achieved 100\% tracking accuracy. As shown in Table. \ref{tab:tab1}, the overall clustering accuracy is over 99\%, hence only a minor refinement was required.

\begin{figure*}[!t]
	\centering
	\subfloat[]{\includegraphics[width=0.2\textwidth]{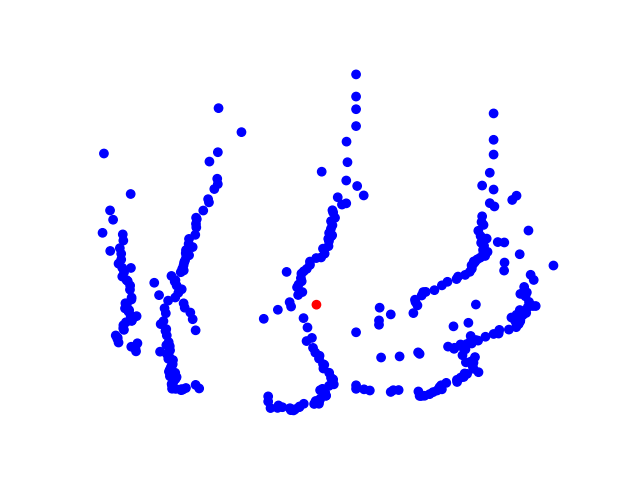}%
		\label{fig:origin}}
	\hfil
	\subfloat[]{\includegraphics[width=0.2\textwidth]{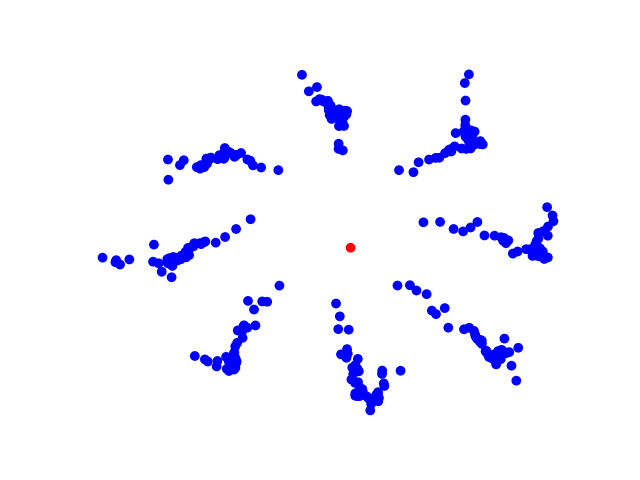}%
		\label{fig:gas}}
	\hfil
	\subfloat[]{\includegraphics[width=0.2\textwidth]{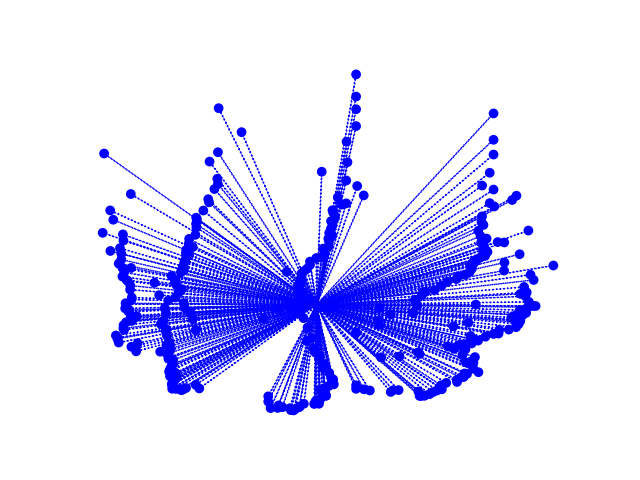}%
		\label{fig:origin_line}}
	\hfil
	\subfloat[]{\includegraphics[width=0.2\textwidth]{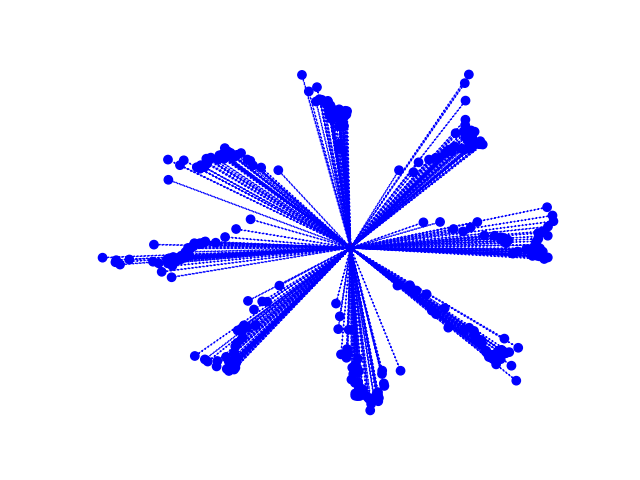}%
		\label{fig:gas_line}}
	\caption{Illustration of the efficiency of GA's fitness function. (a) Nuclei projected to x-z plane. blue: projected nuclei, red: average location of projected nuclei. (b) Nuclei projected to GA selected plane. blue: projected nuclei, red: average location of projected nuclei. (c) Distances from projected nuclei to average location in fig (a). (d) Distances from projected nuclei to average location in fig (b).}
	\label{figFit}
\end{figure*}

\begin{figure}[!t]
	\centering

	\subfloat[]{\includegraphics[width=0.16\textwidth]{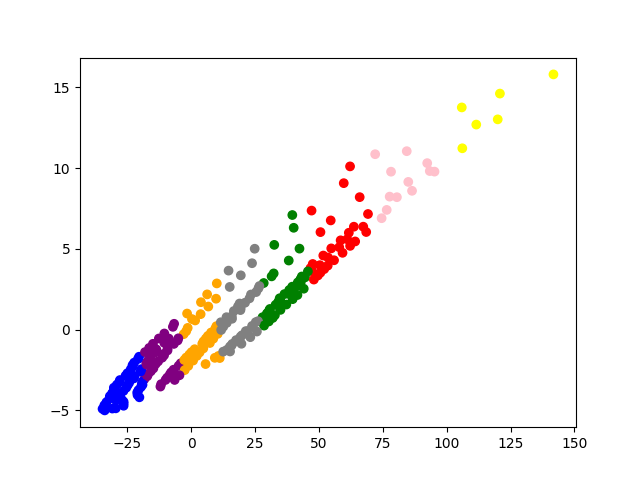}%
		\label{fig:pca1}}
		\hfil
	\subfloat[]{\includegraphics[width=0.16\textwidth]{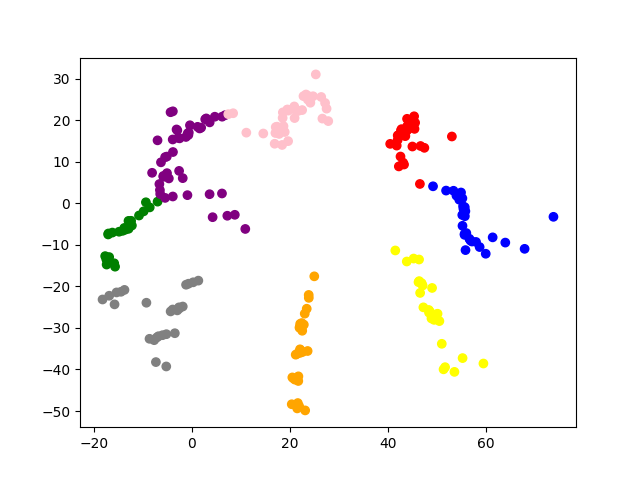}%
		\label{fig:pca2}}
		\hfil
	\subfloat[]{\includegraphics[width=0.16\textwidth]{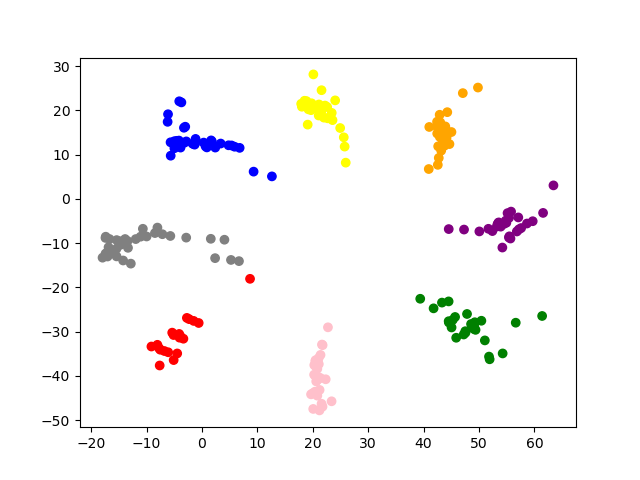}%
		\label{fig:cluster1}}

	\caption{Example of comparison between PCA and proposed GA. (a) The clustreing result on the plane composed by PCA's first component and second component. (b) The clustreing result on the plane composed by PCA's second component and third component. (c) The clustering result based on GA's projection.}
	\label{fig11}
\end{figure}

\hfill \break
Another idea to select a plane for projection is to use principal components analysis (PCA). PCA is used in exploratory data analysis and for making predictive models. It is commonly used for dimensionality reduction by projecting each data point onto a few principal components to obtain lower-dimensional data while preserving as much of data variation as possible. The first principal component can equivalently be defined as a direction that maximizes the variance of the projected data. Accordingly, we tried to project nuclei in a volume onto the plane defined by the first component and the second component, also, we tested on the plane defined by second and third component for comparison. Then, we applied K-means clustering to cluster the projected nuclei into eight groups. As seen in the example result in Fig. \ref{fig11}, this method gave a quantification that was not as accurate as those by our proposed GA method, though it worked to some extent. Quantitative evaluation of the two methods in terms of the accuracy of clustering indicated that our GA-based method outperformed that by PCA as shown in Table \ref{tab:tab1}.\\

\begin{figure}[h]%
	\centering
	\captionsetup{justification=centering}
	\includegraphics[width=0.18\textwidth]{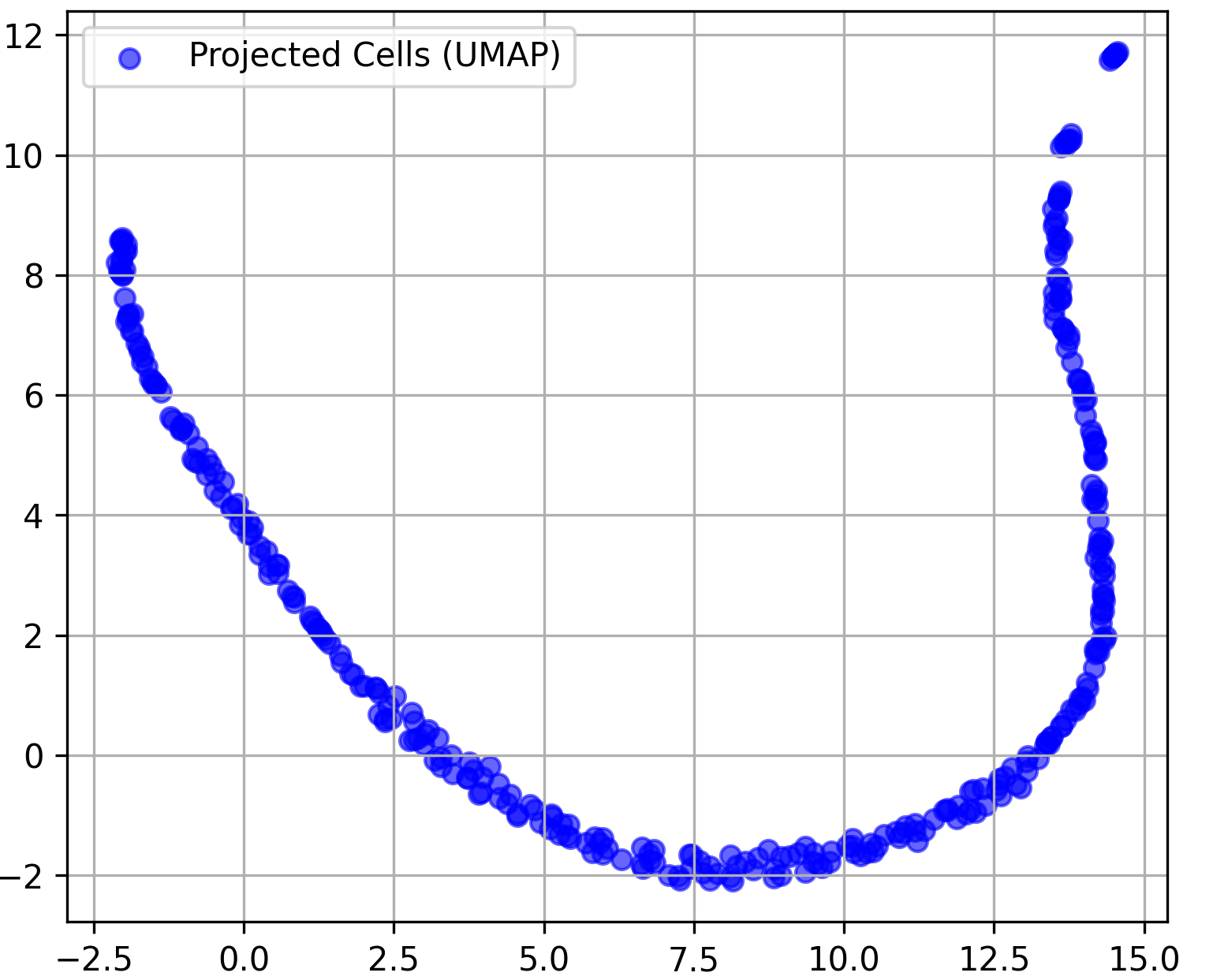}
	\caption{An example of dimension reduction method using UMAP, where the projection does not effectively preserve the original spatial structure of the cells. }
	\label{fig_umap}
\end{figure}

Other dimensionality reduction methods, such as Uniform Manifold Approximation and Projection (UMAP) \cite{mcinnes2018umap}, are effective for capturing nonlinear structures in data. However, these methods do not effectively preserve the original spatial organization of the cells. Specifically, the eight curved line structures are not maintained. As shown in Fig. \ref{fig_umap}, UMAP produces a continuous, distorted embedding in which the individual curved lines become indistinguishable. This result highlights a key limitation of UMAP in our specific task: while UMAP excels at general manifold learning and neighborhood preservation, it does not guarantee a projection that retains the structural integrity of the original 3D spatial distribution. In contrast, our GA-based approach explicitly optimizes a projection plane to ensure a meaningful and biologically relevant visualization. Since UMAP operates as a generic embedding technique without constraints tailored to our data, it does not provide an optimal solution for preserving the spatial arrangement of cell structures.

\section{Discussion} 

Through the experiments described above, our proposed method is proven efficient in tracking Arabidopsis root cortex nuclei in 3D time-lapse imaging datasets. We here make a discussion about our method on several aspects: Fitness function of GA, Robustness of GA and K-means, Necessity of projection (dimension reduction), Other clustering methods(DBSCAN) and Processing time.\\

\subsection{Fitness function of GA}
The most significant component in our method is the use of GA to project the nuclei to the plane most suitable to separate them into eight lines. In the proposed GA, the fitness function decides the quality of a chromosome. In Fig. \ref{figFit}, nuclei are projected either directly to the X-Z plane of original microscopic field (Fig. \ref{fig:origin}) or to the plane selected by GA (Fig. \ref{fig:gas}). Blue dots represent projected nuclei, whereas red dot represents the centroid of the projected nuclei. In Fig. \ref{fig:origin_line} and \ref{fig:gas_line}, each of the projected nuclei was linked to the average locations shown in the respective projection. These representations visualize considerable reduction in the distances when the nuclei are projected to the GA-selected plane. Fitness plays a significant role in GA, where through numbers of iterations, an optimal plane is finally selected. \\

\begin{figure}[!t]
	\centering
	
	\subfloat[]{\includegraphics[width=0.2\textwidth]{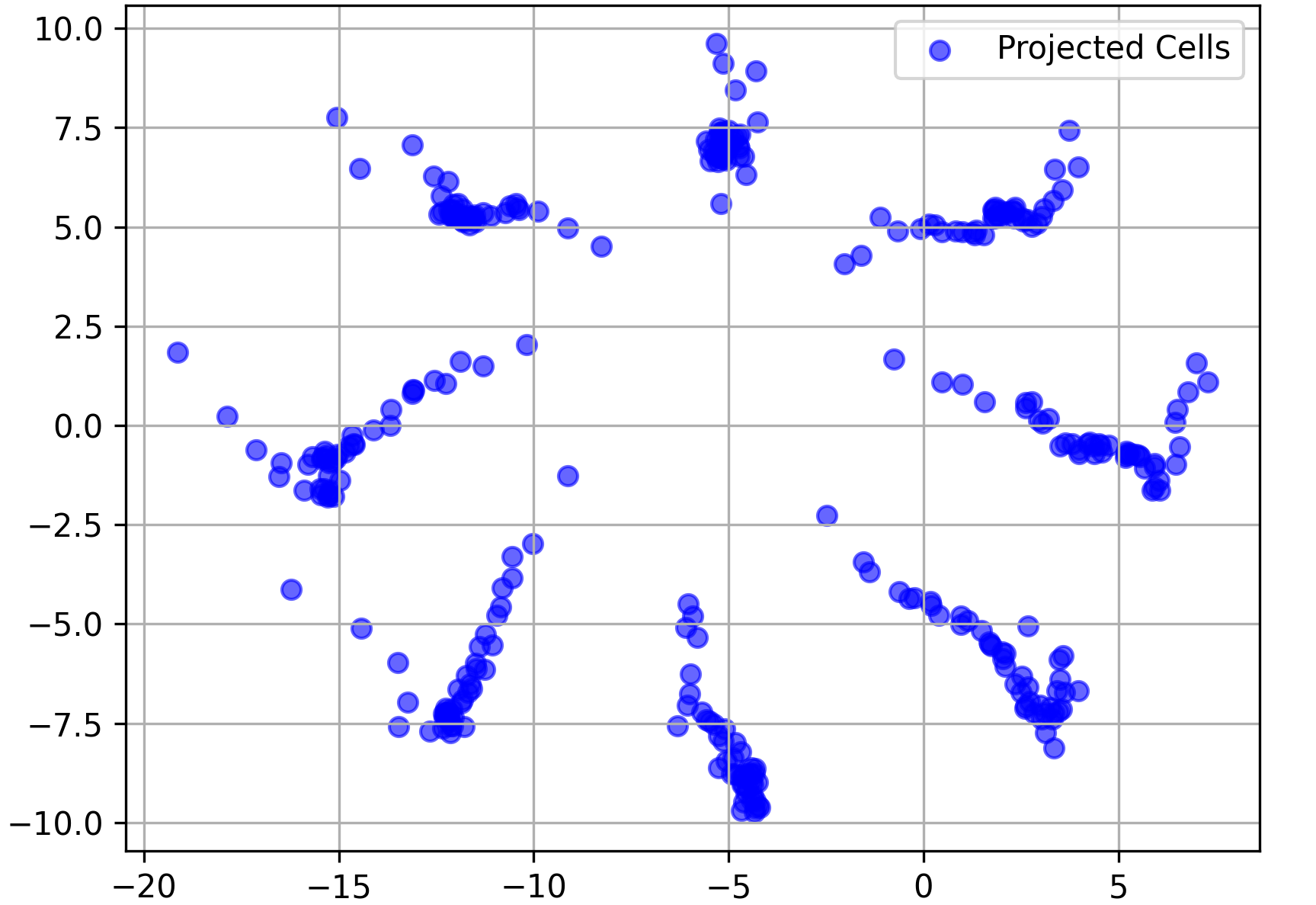}%
		\label{fig:ga_frame2}}
	\hfil
	\subfloat[]{\includegraphics[width=0.2\textwidth]{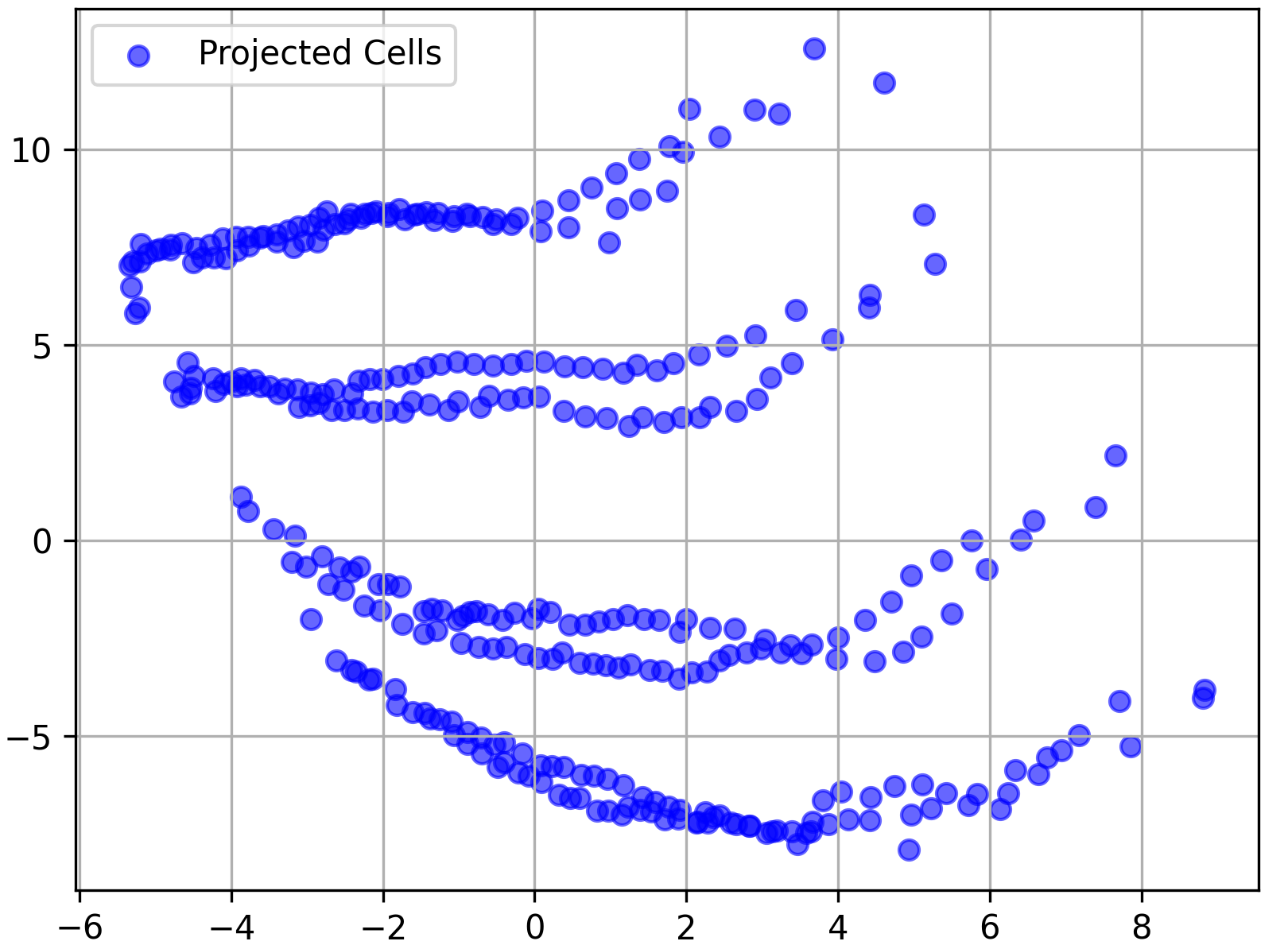}%
		\label{fig:gradient_frame_2}}

	\caption{Example of comparison between L-BFGS-B and proposed GA optimization method. (a) The projection using plane selected by L-BFGS-B (b) The projection using plane selected by L-BFGS-B. }
	\label{fig_gradient}
\end{figure}

Since our proposed fitness function is differentiable, a gradient-based method could potentially be an efficient approach for finding the optimal projection plane. To address this concern, we implemented a gradient-based optimization method, L-BFGS-B \cite{byrd1995limited}, for comparison. We incorporated a constraint to ensure that the projection plane, represented by a unit vector, always maintains a norm of 1. Figure \ref{fig_gradient} presents a visualization comparing GA and L-BFGS-B. Our results indicate that gradient-based optimization often converged to suboptimal local minima, leading to inconsistent projection planes. To further evaluate this behavior, we tested L-BFGS-B across multiple random initializations. We observed that the optimizer did not always converge to the same solution, producing varying fitness values across runs. This inconsistency suggests that gradient-based methods are not ideal for reliably finding the optimal projection plane. Additionally, we conducted a convexity analysis by numerically approximating the Hessian matrix of our fitness function. Using finite difference approximation at multiple points in the optimization space, we found that the Hessian matrix consistently exhibited at least one negative eigenvalue. For instance, at a randomly selected initialization, the eigenvalues were: [0.370,-0.046,0.060]. 
The presence of a negative eigenvalue confirms that the function is not strictly convex, meaning that gradient-based optimizers can become trapped in local minima, leading to suboptimal solutions. In conclusion, while gradient-based methods may perform well under certain conditions, our empirical analysis demonstrates that GA provides a more robust and globally optimal solution for clustering cell lines effectively.\\

Since K-means clustering cannot guarantee 100\% accuracy, we provide manual refinement tools \cite{Goh2022} for correction. In rare cases where two projected points are very close, potentially causing misclustering, these can be easily rectified using our tools. \\

One might question whether a single projection plane is sufficient given the slight curvature of nucleus lines. However, projected nuclei from the same line remain close enough for effective clustering. As shown in Table \ref{tab:tab1}, our method achieves 99.53\% accuracy, requiring only minor corrections. Using two projection planes would double computation time with minimal accuracy improvement, making it inefficient.

\subsection{Robustness of GA and K-means}
\begin{figure}[!t]
	\centering
	\subfloat[rotate -0.2 $^{\circ}$]{\includegraphics[width=0.16\textwidth]{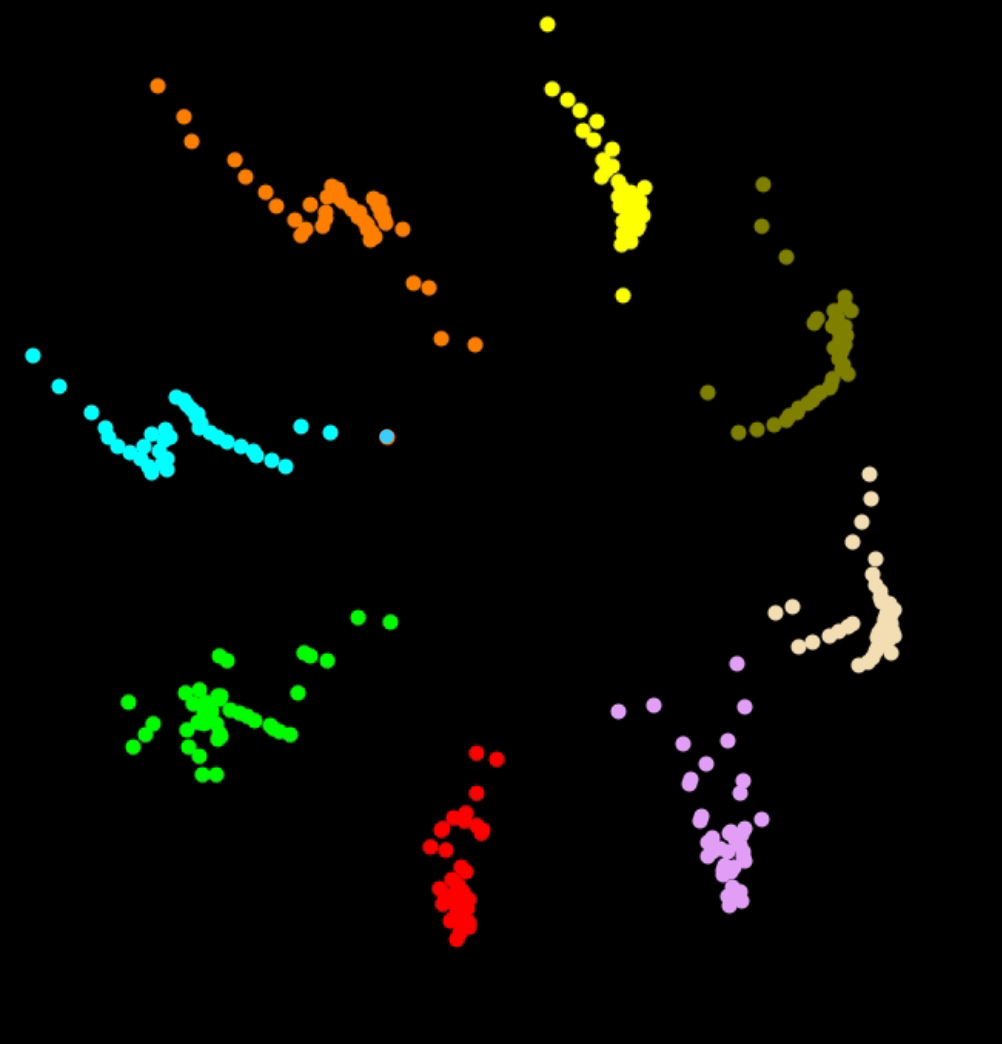}%
		\label{fig:xyz-0.2}}
	\hfil
	\subfloat[rotate -0.1 $^{\circ}$]{\includegraphics[width=0.16\textwidth]{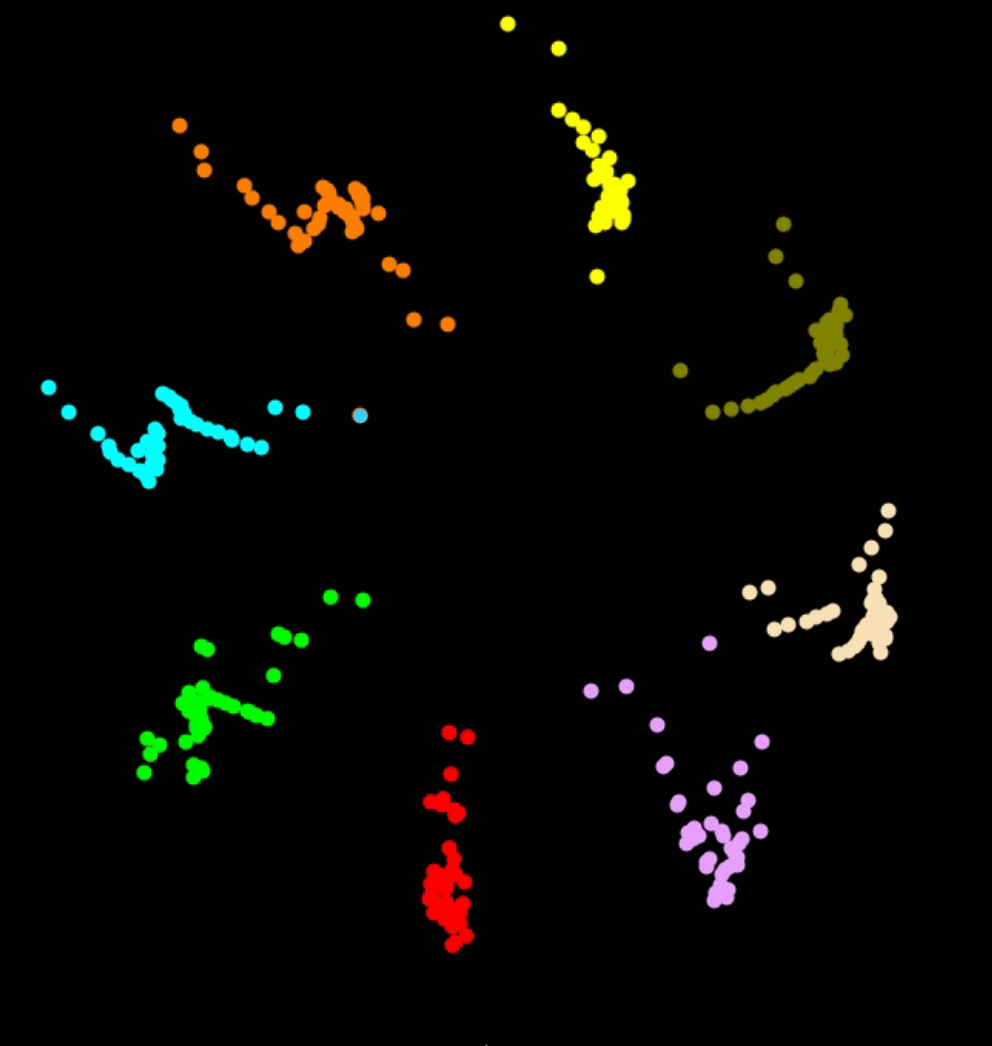}%
		\label{fig:xyz-0.1}}
	\hfil
	\subfloat[original plane]{\includegraphics[width=0.16\textwidth]{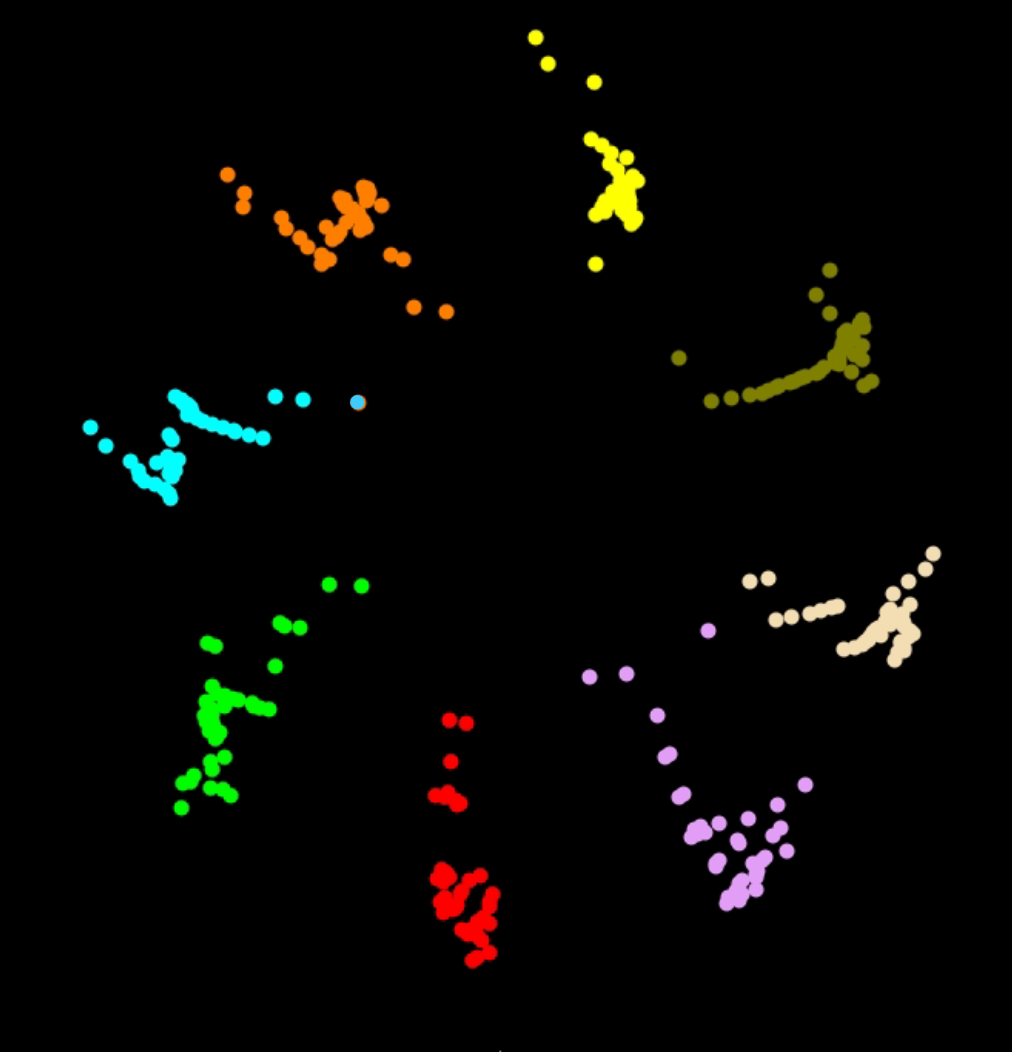}%
		\label{fig:xyz0}}
	\hfil
	\subfloat[rotate 0.1 $^{\circ}$]{\includegraphics[width=0.16\textwidth]{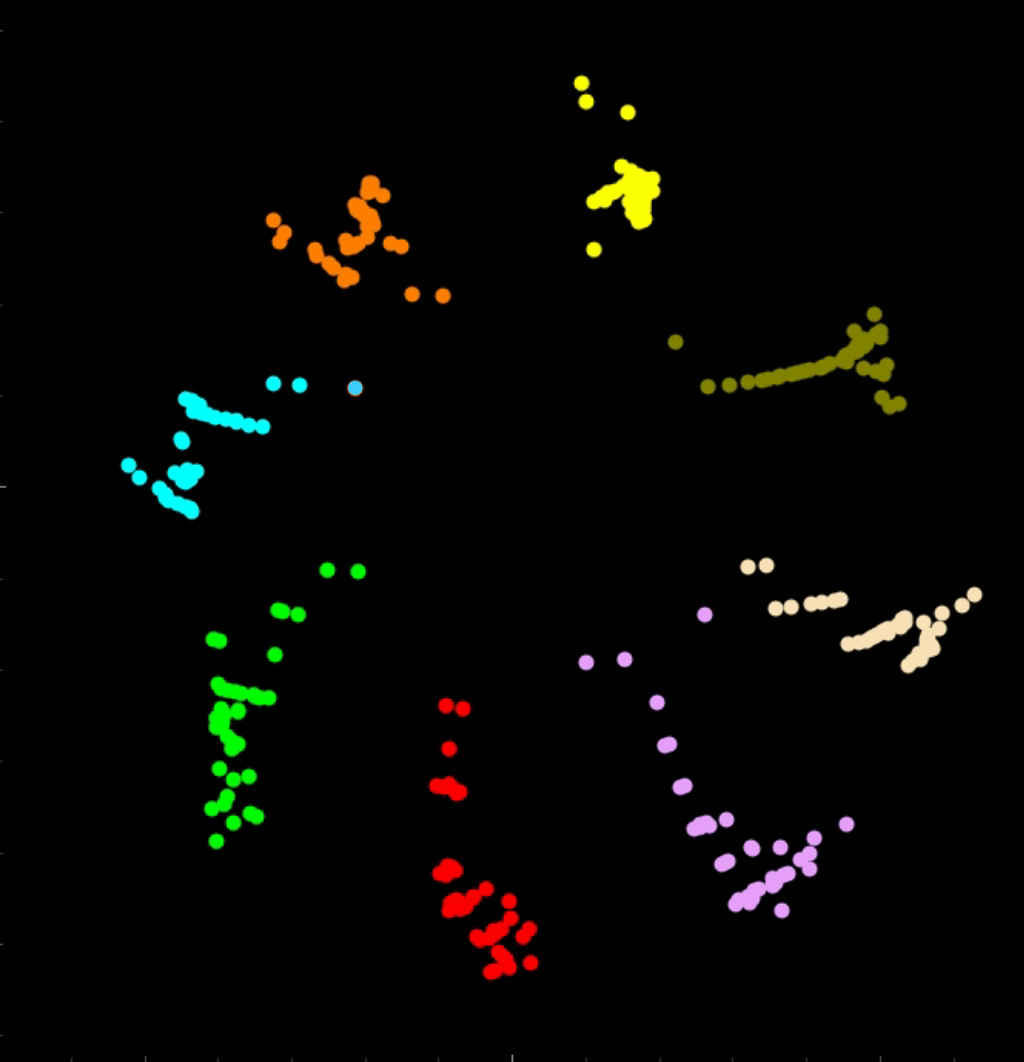}%
		\label{fig:xyz0.1}}
	\hfil
		\subfloat[rotate 0.2 $^{\circ}$]{\includegraphics[width=0.16\textwidth]{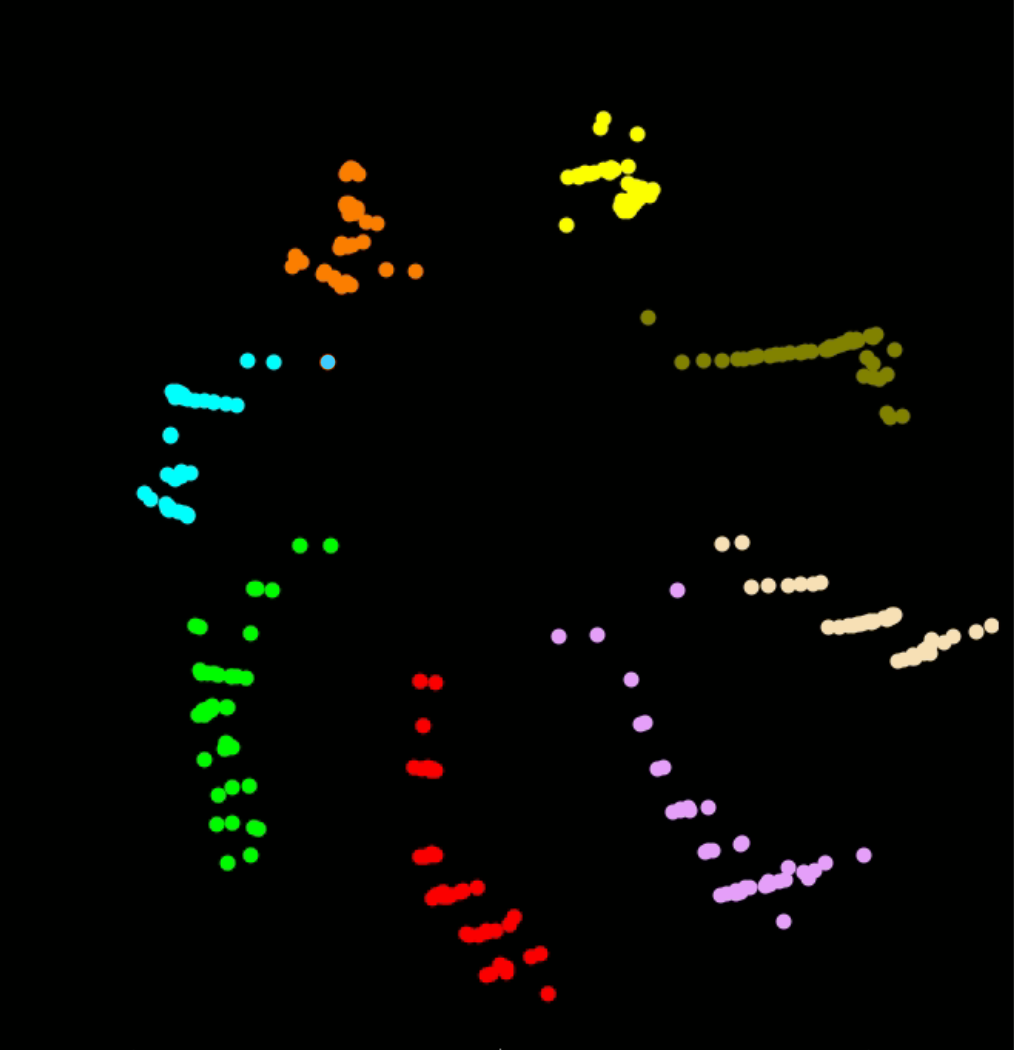}%
		\label{fig:xyz0.2}}
	\caption{Rotation of a plane ranges from -0.2 $^{\circ}$ to 0.2 $^{\circ}$, where the step is 0.1 $^{\circ}$. The original plane represents no rotation.}
	\label{rotation}
\end{figure}

Genetic Algorithms (GAs) can produce slight variations across runs, raising concerns about robustness. To evaluate this, we ran GA 100 times on the same dataset. The largest observed variance resulted in a plane rotation of no more than 0.1 degrees in any direction (x, y, z), which had minimal impact on results. Figure \ref{rotation} illustrates this negligible effect with rotations ranging from -0.2 to 0.2 degrees.  Despite minor variations, GA consistently converges to a globally optimal solution by efficiently exploring the search space.

K-means, though sensitive to initial conditions, consistently achieved high clustering accuracy when combined with GA. Across 100 runs, the accuracy remained around 99.5\%, with a maximum variation of only 0.1\%. These findings confirm the robustness of our approach.

\subsection{K-means VS DBSCAN}
Clustering is essential in our method, and selecting an approach that accurately separates the eight cell files is critical. DBSCAN, known for its flexibility in detecting clusters of varying shapes and densities, is robust to outliers by classifying noise points. However, these strengths become drawbacks in our task.  

Since Arabidopsis root cortex cells consistently form eight files, K-means, which explicitly partitions data into a set number of clusters, is more suitable. In contrast, DBSCAN does not guarantee eight clusters, requiring additional corrections. Moreover, DBSCAN’s sensitivity to parameter $\epsilon$ makes it difficult to handle distant nuclei within the same file, potentially misclassifying them as noise.  

\begin{figure}[!t]
	\centering
	\subfloat[]{\includegraphics[width=0.2\textwidth]{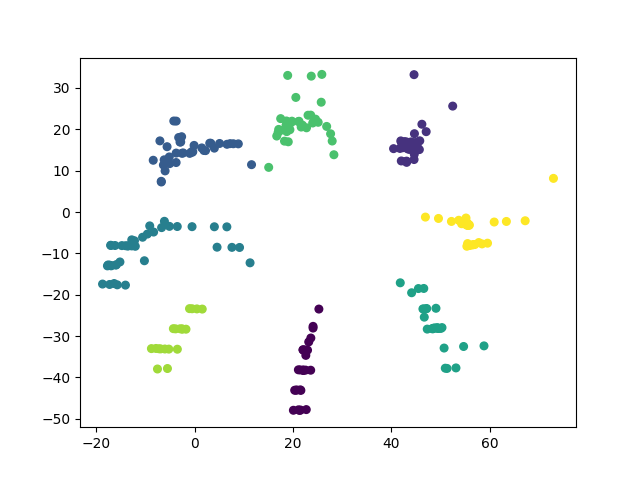}%
		\label{fig:k_means}}
	\hfil
	\subfloat[]{\includegraphics[width=0.2\textwidth]{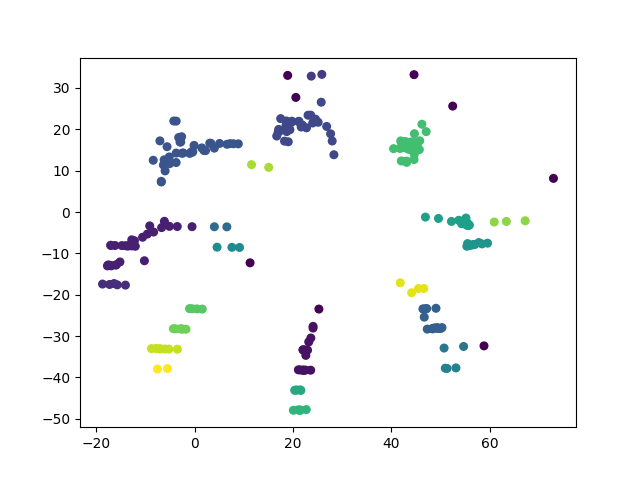}%
		\label{fig:dbscan}}
	\caption{Example of comparison between K-means and DBSCAN, where same color represents same cluster. (a) The clustering result based on K-means. (b) The clustreing result based on DBSCAN.}
	\label{fig12}
\end{figure}

As shown in Figure \ref{fig12}, K-means achieves more reliable clustering for our specific dataset.

\subsection{Necessity of Projection}
While K-means can be applied to 3D datasets, it's important to address the necessity of projection in our approach. In a 3D scenario, the primary method for clustering would typically rely on the distances between nuclei, as the visual appearance of nuclei may lack distinctive features. However, the distances between cells can be extremely close, making K-means less effective. Through experiments, we have determined that K-means does not perform well in a 3D scenario. Consequently, we propose the projection of nuclei onto a plane, which enhances the distinguishability of different lines of nuclei. \\

\subsection{Cell Tracking vs Multi-Object Tracking}
Cell tracking in microscopy videos shares similarities with the widely explored multiple object tracking (MOT) problem in computer vision. However, there are two key distinctions that set cell tracking apart and drive the need for specialized approaches. First, microscopy images are typically grayscale and contain thousands of visually indistinguishable cells, making local differentiation highly challenging. Second, unlike objects in conventional MOT tasks, cells can divide but generally do not merge, which significantly increases the complexity of the tracking problem. This introduces a combinatorial challenge that renders many standard MOT techniques, such as those based on similarity matching \cite{suljagic2022similarity}, ineffective.

To investigate this limitation, we conducted experiments using the method proposed in \cite{suljagic2022similarity}, applying it to detected cells across consecutive frames. Our analysis revealed that every cell in frame $t$ exhibited similarity scores exceeding 0.9 when compared to multiple cells in frame 
$t+1$, leading to severe ambiguity. This result demonstrates that MOT methods, particularly those relying on similarity-based matching, struggle to handle the unique challenges of cell tracking and are unsuitable for accurately tracking individual cells in dense biological environments.

\subsection{Processing Time}
Our proposed method places a strong emphasis on achieving precise tracking of Arabidopsis root cortex cells, prioritizing accuracy over processing speed. To enhance the efficiency of our GA method, we have incorporated multi-threading into our system, as detailed in \cite{Goh2022}.

\begin{table}[!t]
	\caption{Performance comparison between PCA and GA on 90 frames\label{tab:tab1}}
	\centering
	\begin{tabular}{@{}lll@{}}
		\hline   & Clustering Accuracy \\
		\hline
		PCA($1^{st}$ and $2^{nd}$ component)   & 40.63\%   \\
		PCA($2^{nd}$ and $3^{rd}$ component)   & 94.21\%   \\
		GA    & 99.53\%  \\
		\hline
	\end{tabular}
	\footnotetext{The comparison of clustering accuracy between GA and PCA}
\end{table}

\section{Conclusion}
In this paper, we proposed a GA-based method for tracking Arabidopsis root cortex nuclei by leveraging their structured 3D arrangement. Our approach optimizes a projection plane using GA, clusters nuclei into eight groups with K-means, and tracks them across frames using polar coordinates. It achieved 100\% tracking accuracy with minor manual correction.

Furthermore, we are developing an automatic nuclear detection method \cite{Goh2022, song2024dividing}, which, when integrated with our tracking approach, enables a tracking-by-detection framework for analyzing Arabidopsis root's cellular dynamics.

Our system, implemented in Python with PySide2, is publicly available at  \url{https://github.com/JerrySongCST/Arabidopsis_root_cortex_cell_tracking}.\\

\section*{Acknowledgments}
This work was partially supported by Grant-in-Aid for Scientific Research on Innovative Areas "Periodicity and Its Modulation in Plants" from the Ministry of Education, Culture, Sports, Science and Technology, Japan, 20H05428 and 22H04736 to YWC, 19H05670 to KN and YK, 19H05671 to KN, a grant from Japan Science and Technology Agency (JST) PRESTO "Multicellular System" to TG, and the JST SPRING JPMJSP2101 to YS.


\vfill


\begin{thebibliography}{1} 
\bibliographystyle{IEEEtran}

\bibitem{van1995cell} Berg, C., Willemsen, V., Hage, W., Weisbeek, P. \& Scheres, B. Cell fate in the Arabidopsis root meristem determined by directional signalling. {\em Nature}. \textbf{378}, 62-65 (1995)

\bibitem{sabatini1999auxin} Sabatini, S., Beis, D., Wolkenfelt, H., Murfett, J., Guilfoyle, T., Malamy, J., Benfey, P., Leyser, O., Bechtold, N., Weisbeek, P. \& Others An auxin-dependent distal organizer of pattern and polarity in the Arabidopsis root. {\em Cell}. \textbf{99}, 463-472 (1999)

\bibitem{eldar2010functional} Eldar, A. \& Elowitz, M. Functional roles for noise in genetic circuits. {\em Nature}. \textbf{467}, 167-173 (2010)

\bibitem{moreno2010oscillating} Moreno-Risueno, M., Van Norman, J., Moreno, A., Zhang, J., Ahnert, S. \& Benfey, P. Oscillating gene expression determines competence for periodic Arabidopsis root branching. {\em Science}. \textbf{329}, 1306-1311 (2010)

\bibitem{heisler2005patterns}Heisler, M., Ohno, C., Das, P., Sieber, P., Reddy, G., Long, J. \& Meyerowitz, E. Patterns of auxin transport and gene expression during primordium development revealed by live imaging of the Arabidopsis inflorescence meristem. {\em Current Biology}. \textbf{15}, 1899-1911 (2005)

\bibitem{jonsson2006auxin}Jönsson, H., Heisler, M., Shapiro, B., Meyerowitz, E. \& Mjolsness, E. An auxin-driven polarized transport model for phyllotaxis. {\em Proceedings Of The National Academy Of Sciences}. \textbf{103}, 1633-1638 (2006)

\bibitem{bragantini2024large}Bragantini, J., Lange, M. \& Royer, L. Large-scale multi-hypotheses cell tracking using ultrametric contours maps. {\em European Conference On Computer Vision}. pp. 36-54 (2024)

\bibitem{jug2016moral}Jug, F., Levinkov, E., Blasse, C., Myers, E. \& Andres, B. Moral lineage tracing. {\em Proceedings Of The IEEE Conference On Computer Vision And Pattern Recognition}. pp. 5926-5935 (2016)

\bibitem{Goh2022}Goh, T., Song, Y., Yonekura, T., Obushi, N., Den, Z., Imizu, K., Tomizawa, Y., Kondo, Y., Miyashima, S., Iwamoto, Y. \& Others In-Depth Quantification of Cell Division and Elongation Dynamics at the Tip of Growing Arabidopsis Roots Using 4D Microscopy, AI-Assisted Image Processing and Data Sonification. {\em Plant And Cell Physiology}. pp. pcad105 (2023).

\bibitem{rahni2019week}Rahni, R. \& Birnbaum, K. Week-long imaging of cell divisions in the Arabidopsis root meristem. {\em Plant Methods}. \textbf{15}, 1-14 (2019)

\bibitem{von2017live}Von Wangenheim, D., Hauschild, R., Fendrych, M., Barone, V., Benkova, E. \& Friml, J. Live tracking of moving samples in confocal microscopy for vertically grown roots. {\em Elife}. \textbf{6} pp. e26792 (2017)

\bibitem{jaqaman2008robust}Jaqaman, K., Loerke, D., Mettlen, M., Kuwata, H., Grinstein, S., Schmid, S. \& Danuser, G. Robust single-particle tracking in live-cell time-lapse sequences. {\em Nature Methods}. \textbf{5}, 695-702 (2008)

\bibitem{munkres1957algorithms}Munkres, J. Algorithms for the assignment and transportation problems. {\em Journal Of The Society For Industrial And Applied Mathematics}. \textbf{5}, 32-38 (1957)

\bibitem{mitchell1998introduction}Mitchell, M. An introduction to genetic algorithms. (MIT press,1998)

\bibitem{janes2018cellular}Janes, G., Von Wangenheim, D., Cowling, S., Kerr, I., Band, L., French, A. \& Bishopp, A. Cellular patterning of Arabidopsis roots under low phosphate conditions. {\em Frontiers In Plant Science}. \textbf{9} pp. 735 (2018)

\bibitem{whitley1994genetic}Whitley, D. A genetic algorithm tutorial. {\em Statistics And Computing}. \textbf{4}, 65-85 (1994)

\bibitem{metzler2014anomalous}Metzler, R., Jeon, J., Cherstvy, A. \& Barkai, E. Anomalous diffusion models and their properties: non-stationarity, non-ergodicity, and ageing at the centenary of single particle tracking. {\em Physical Chemistry Chemical Physics}. \textbf{16}, 24128-24164 (2014)

\bibitem{manzo2015review}Manzo, C. \& Garcia-Parajo, M. A review of progress in single particle tracking: from methods to biophysical insights. {\em Reports On Progress In Physics}. \textbf{78}, 124601 (2015)

\bibitem{anthony2006methods}Anthony, S., Zhang, L. \& Granick, S. Methods to track single-molecule trajectories. {\em Langmuir}. \textbf{22}, 5266-5272 (2006)

\bibitem{lindeberg2013scale}Lindeberg, T. Scale selection properties of generalized scale-space interest point detectors. {\em Journal Of Mathematical Imaging And Vision}. \textbf{46}, 177-210 (2013)

\bibitem{macqueen1967some}MacQueen, J. \& Others Some methods for classification and analysis of multivariate observations. {\em Proceedings Of The Fifth Berkeley Symposium On Mathematical Statistics And Probability}. \textbf{1}, 281-297 (1967)

\bibitem{schneider2012nih}Schneider, C., Rasband, W. \& Eliceiri, K. NIH Image to ImageJ: 25 years of image analysis. {\em Nature Methods}. \textbf{9}, 671-675 (2012)

\bibitem{lindeberg1998feature}Lindeberg, T. Feature detection with automatic scale selection. {\em International Journal Of Computer Vision}. \textbf{30}, 79-116 (1998)

\bibitem{ren2015faster}Ren, S., He, K., Girshick, R. \& Sun, J. Faster r-cnn: Towards real-time object detection with region proposal networks. {\em Advances In Neural Information Processing Systems}. \textbf{28} (2015)

\bibitem{redmon2018yolov3}Redmon, J. \& Farhadi, A. Yolov3: An incremental improvement. {\em ArXiv Preprint ArXiv:1804.02767}. (2018)

\bibitem{zhou2019objects}Zhou, X., Wang, D. \& Krähenbühl, P. Objects as points. {\em ArXiv Preprint ArXiv:1904.07850}. (2019)

\bibitem{carion2020end}Carion, N., Massa, F., Synnaeve, G., Usunier, N., Kirillov, A. \& Zagoruyko, S. End-to-end object detection with transformers. {\em European Conference On Computer Vision}. pp. 213-229 (2020)

\bibitem{bewley2016simple}Bewley, A., Ge, Z., Ott, L., Ramos, F. \& Upcroft, B. Simple online and realtime tracking. {\em 2016 IEEE International Conference On Image Processing (ICIP)}. pp. 3464-3468 (2016)

\bibitem{wang2019ranet}Wang, Z., Xu, J., Liu, L., Zhu, F. \& Shao, L. Ranet: Ranking attention network for fast video object segmentation. {\em Proceedings Of The IEEE/CVF International Conference On Computer Vision}. pp. 3978-3987 (2019)

\bibitem{milan2017online}Milan, A., Rezatofighi, S., Dick, A., Reid, I. \& Schindler, K. Online multi-target tracking using recurrent neural networks. {\em Thirty-First AAAI Conference On Artificial Intelligence}. (2017)

\bibitem{kim2018multi}Kim, C., Li, F. \& Rehg, J. Multi-object tracking with neural gating using bilinear lstm. {\em Proceedings Of The European Conference On Computer Vision (ECCV)}. pp. 200-215 (2018)

\bibitem{zhang2021fairmot}Zhang, Y., Wang, C., Wang, X., Zeng, W. \& Liu, W. Fairmot: On the fairness of detection and re-identification in multiple object tracking. {\em International Journal Of Computer Vision}. \textbf{129}, 3069-3087 (2021)

\bibitem{meinhardt2022trackformer}Meinhardt, T., Kirillov, A., Leal-Taixe, L. \& Feichtenhofer, C. Trackformer: Multi-object tracking with transformers. {\em Proceedings Of The IEEE/CVF Conference On Computer Vision And Pattern Recognition}. pp. 8844-8854 (2022)

\bibitem{sun2020transtrack}Sun, P., Cao, J., Jiang, Y., Zhang, R., Xie, E., Yuan, Z., Wang, C. \& Luo, P. Transtrack: Multiple object tracking with transformer. {\em ArXiv Preprint ArXiv:2012.15460}. (2020)

\bibitem{bergmann2019tracking}Bergmann, P., Meinhardt, T. \& Leal-Taixe, L. Tracking without bells and whistles. {\em Proceedings Of The IEEE/CVF International Conference On Computer Vision}. pp. 941-951 (2019)

\bibitem{chu2019famnet}Chu, P. \& Ling, H. Famnet: Joint learning of feature, affinity and multi-dimensional assignment for online multiple object tracking. {\em Proceedings Of The IEEE/CVF International Conference On Computer Vision}. pp. 6172-6181 (2019)

\bibitem{feichtenhofer2017detect}Feichtenhofer, C., Pinz, A. \& Zisserman, A. Detect to track and track to detect. {\em Proceedings Of The IEEE International Conference On Computer Vision}. pp. 3038-3046 (2017)

\bibitem{hayashida2020mpm}Hayashida, J., Nishimura, K. \& Bise, R. MPM: Joint representation of motion and position map for cell tracking. {\em Proceedings Of The IEEE/CVF Conference On Computer Vision And Pattern Recognition}. pp. 3823-3832 (2020)

\bibitem{chen2002estimating}Chen, Y., Mendoza, N., Nakao, Z. \& Adachi, T. Estimating wind speed in the lower atmosphere wind profiler based on a genetic algorithm. {\em IEEE Transactions On Instrumentation And Measurement}. \textbf{51}, 593-597 (2002)

\bibitem{zhu2022optimizing}Zhu, J., Zhu, C., Zhou, P., He, Z. \& You, D. An Optimizing Diffusion Kernel-Based Binary Encoding Strategy With Genetic Algorithm for Fringe Projection Profilometry. {\em IEEE Transactions On Instrumentation And Measurement}. \textbf{71} pp. 1-8 (2022)

\bibitem{capelli2017genetic}Capelli, F., Riba, J., Ruperez, E. \& Sanllehi, J. A genetic-algorithm-optimized fractal model to predict the constriction resistance from surface roughness measurements. {\em IEEE Transactions On Instrumentation And Measurement}. \textbf{66}, 2437-2447 (2017)

\bibitem{olmi2000genetic}Olmi, R., Bini, M. \& Priori, S. A genetic algorithm approach to image reconstruction in electrical impedance tomography. {\em IEEE Transactions On Evolutionary Computation}. \textbf{4}, 83-88 (2000)
2000, pp. 2245--2249.

\bibitem{ester1996density}Ester, M., Kriegel, H., Sander, J., Xu, X. \& Others A density-based algorithm for discovering clusters in large spatial databases with noise.. {\em Kdd}. \textbf{96}, 226-231 (1996)

\bibitem{RonnebergerFB15}Ronneberger, O., Fischer, P. \& Brox, T. U-Net: Convolutional Networks for Biomedical Image Segmentation. {\em CoRR}. \textbf{abs/1505.04597} (2015), http://arxiv.org/abs/1505.04597

\bibitem{mcinnes2018umap}McInnes, L., Healy, J. \& Melville, J. Umap: Uniform manifold approximation and projection for dimension reduction. {\em ArXiv Preprint ArXiv:1802.03426}. (2018)

\bibitem{byrd1995limited}Byrd, R., Lu, P., Nocedal, J. \& Zhu, C. A limited memory algorithm for bound constrained optimization. {\em SIAM Journal On Scientific Computing}. \textbf{16}, 1190-1208 (1995)

\bibitem{suljagic2022similarity}Suljagic, H., Bayraktar, E. \& Celebi, N. Similarity based person re-identification for multi-object tracking using deep Siamese network. {\em Neural Computing And Applications}. \textbf{34}, 18171-18182 (2022)

\bibitem{song2024dividing}Song, Y., Deng, Z., Li, Y. \& Others Dividing and Non-dividing Cell Detection by Segmentation on Arabidopsis Root Images Using Light-weight U-Net. {\em IEICE Proceedings Series}. \textbf{81} (2024)


\end{thebibliography}
\end{document}